\documentclass[10pt,journal,compsoc]{IEEEtran}

\ifCLASSOPTIONcompsoc
  \usepackage[nocompress]{cite}
\else
  \usepackage{cite}
\fi

\usepackage{amsmath}
\usepackage{mathtools}
\usepackage{amssymb}
\usepackage{amsthm}
\usepackage{siunitx}
\usepackage{physics}
\usepackage{bm}
\usepackage{algorithm}
\usepackage{algcompatible}
\usepackage{soul}
\usepackage{graphicx}
\usepackage{xcolor}\usepackage{amsmath}
\usepackage{mathtools}
\usepackage{amssymb}
\usepackage{amsthm}
\usepackage{siunitx}
\usepackage{physics}
\usepackage{bm}
\usepackage{algorithm}
\usepackage{algcompatible}
\usepackage{soul}
\usepackage{graphicx}
\usepackage{xcolor}
\usepackage{tabularx}
\usepackage{amsfonts,multirow,bigstrut,booktabs,ctable,latexsym}
\AtBeginDocument{\RenewCommandCopy\qty\SI}
\ExplSyntaxOn
\msg_redirect_name:nnn { siunitx } { physics-pkg } { none }
\ExplSyntaxOff

\makeatletter

\newtheorem{definition}{\textbf{Definition}}
\newtheorem{lemma}{\textbf{Lemma}}
\newtheorem{theorem}{\textbf{Theorem}}

\newtheorem{assumption}{\textbf{Assumption}}
\theoremstyle{definition}

\newcounter{relctr} 
\everydisplay\expandafter{\the\everydisplay\setcounter{relctr}{0}} 

\newcommand\labelrel[2]{%
  \begingroup
    \refstepcounter{relctr}%
    \stackrel{\textnormal{(\alph{relctr})}}{\mathstrut{#1}}%
    \originallabel{#2}%
  \endgroup
}
\AtBeginDocument{\let\originallabel\label} 

\ifCLASSINFOpdf
\else
\fi
\DeclareMathOperator{\randk}{rand} 

\ifCLASSOPTIONcompsoc
 \usepackage[caption=false,font=footnotesize,labelfont=sf,textfont=sf]{subfig}
\else
 \usepackage[caption=false,font=footnotesize]{subfig}
\fi

\begin{document}
\bstctlcite{bstctl:nodash}
\title{Communication and Energy Efficient Wireless Federated Learning with Intrinsic Privacy}
\author{Zhenxiao~Zhang,~\IEEEmembership{Student Member,~IEEE,} Yuanxiong~Guo,~\IEEEmembership{Senior Member,~IEEE,} Yuguang~Fang,~\IEEEmembership{Fellow,~IEEE,}         and~Yanmin~Gong,~\IEEEmembership{Senior~Fellow,~IEEE}
\IEEEcompsocitemizethanks{\IEEEcompsocthanksitem Z. Zhang, Y. Guo, and Y. Gong are with the University of Texas at San Antonio, San Antonio, TX USA 78249 (e-mail: zhenxiao.zhang@utsa.edu, yuanxiong.guo@utsa.edu yanmin.gong@utsa.edu). 
\IEEEcompsocthanksitem Y. Fang is with the City University of Hong Kong, Kowloon Tong, Hong Kong (e-mail: my.fang@cityu.edu.hk). 
}
}

\markboth{}%
{Zhang \MakeLowercase{\textit{et al.}}: Communication and Energy Efficient Wireless Federated Learning with Intrinsic Privacy}

\IEEEtitleabstractindextext{%
\begin{abstract}
Federated Learning (FL) is a collaborative learning framework that enables edge devices to collaboratively learn a global model while keeping raw data locally. Although FL avoids leaking direct information from local datasets, sensitive information can still be inferred from the shared models. To address the privacy issue in FL, differential privacy (DP) mechanisms are leveraged to provide formal privacy guarantee. However, when deploying FL at the wireless edge with over-the-air computation, ensuring client-level DP faces significant challenges. In this paper, we propose a novel wireless FL scheme called private federated edge learning with sparsification (PFELS) to provide client-level DP guarantee with intrinsic channel noise while reducing communication and energy overhead and improving model accuracy. The key idea of PFELS is for each device to first compress its model update and then adaptively design the transmit power of the compressed model update according to the wireless channel status without any artificial noise addition. We provide a privacy analysis for PFELS and prove the convergence of PFELS under general non-convex and non-IID settings. Experimental results show that compared with prior work, PFELS can improve the accuracy with the same DP guarantee and save communication and energy costs simultaneously. 
\end{abstract}

}

\maketitle

\IEEEdisplaynontitleabstractindextext
\IEEEpeerreviewmaketitle

\IEEEraisesectionheading{\section{Introduction}\label{sec:introduction}}

\IEEEPARstart {F}{ederated} Learning (FL) is a distributed machine learning (ML) paradigm in which edge devices collaboratively learn a shared model under the orchestration of a central server while keeping data locally\cite{mcmahan2017communication}. It has gained significant attention because of its inherent privacy preservation and higher efficiency in comparison with conventional centralized ML, which relies heavily on a trustworthy and powerful central server for model training. Specifically, in each communication round, FL consists of four basic stages: (i) the central server sends the current global model to the selected edge devices; (ii) edge devices update their models using the local training data; (iii) edge devices transmit their model updates back to the central server; and (iv) the central server aggregates devices' updates into a new global model. This process is repeated for multiple communication rounds until the global model converges with a satisfactory accuracy. Although FL only requires the transmission of model updates between edge devices and the server instead of raw data, such model transfer can become a communication bottleneck, especially when dealing with modern deep neural networks (DNNs) that has a huge number of parameters (e.g., on the order of hundreds MB, or even GB) \cite{shi2022towards,guo2022hybrid}. Additionally, the transmit power of edge devices is often limited in FL. Thus, it is crucial to design transmission protocols that are both communication and energy-efficient in FL.

When deploying FL over wireless edge, wireless FL \cite{zeng2021energy} where edge devices under the coverage of a nearby access point are coordinated by its co-located edge server to perform FL has been proposed. In wireless FL, the uplink transmissions for model uploading are particularly challenging due to the shared nature of wireless medium among all participating devices. Orthogonal multiple access techniques such as time-division multiple access (TDMA), code-division multiple access (CDMA), and orthogonal frequency-division multiple access (OFDMA) are often used to share the wireless spectrum among the devices. However, as the number of edge devices in wireless FL increases, the spectrum resource that can be allocated to each device decreases proportionally, leading to high latency and low quality of model update transmissions from edge devices to the server. Therefore, enabling scalable FL over wireless edge with a large number of edge devices and limited spectrum resource is very challenging.

Besides the challenge associated with communication, privacy is another core challenge in wireless FL. Although edge devices in wireless FL keep their data locally and only exchange ephemeral model updates that contain less information than raw data, this is not sufficient to guarantee data privacy. For example, by observing the model updates from an edge device, it is possible for the adversary to recover the private dataset in that device using reconstruction attack \cite{fredrikson2015model} or infer whether a sample is in the dataset of that device using membership inference attack \cite{salem2019ml}. Specifically, if the server is not fully trusted, it can easily infer the private information of edge devices from the received model updates during the training by employing existing attack methods. Therefore, how to defend against those advanced privacy attacks and provide rigorous privacy guarantee for each device in wireless FL without a fully trusted edge server is challenging and needs to be addressed. As the state-of-the-art privacy notion, differential privacy (DP) \cite{dwork2014algorithmic} can ensure formal and rigorous privacy protection in FL by adding random noise to the shared model updates. However, the existing DP mechanisms could severely degrade the model accuracy in FL \cite{McMahan2018learning}. How to achieve a strong DP guarantee while maintaining high model accuracy is still challenging in wireless FL.

Over-the-air computation (AirComp) provides a promising solution to addressing both of the aforementioned spectrum and privacy challenges in an integrated manner by achieving scalable and bandwidth-efficient model update aggregation in wireless FL \cite{yang2020federated}. The basic idea of AirComp is to create and leverage the inter-user interference in the multiple access channel (MAC) to boost throughput. In applying AirComp to wireless FL, devices send their model updates in an uncoded manner by directly mapping each model update parameter to a channel symbol: each device first precodes the transmitted symbols by the inverse of the uplink channel gain and then transmits the precoded symbols to the edge server in an analog manner. All the participating devices transmit simultaneously on the same channel so that their signals are aligned and decoded to obtain desired arithmetic computation results at the edge server. In comparison with the traditional orthogonal multiple access techniques where computing and communications are separately done, AirComp is a joint compute-and-communicate scheme by exploiting the fact that MAC inherently yields an additive superposed signal. Note that by using AirComp in wireless FL, the uplink transmission rate of each device does not degrade as the number of devices increases, making it more scalable. Furthermore, the superposition property of wireless channels in AirComp provides an additional benefit to privacy protection. Specifically, by using AirComp in wireless FL, the edge server will only receive from all participating devices the superposition of transmitted signals computed from their private datasets. From the privacy perspective, this makes it harder for the edge server to infer each device's private dataset since the edge server does not know the transmitted signal from each individual device. Moreover, the channel noise naturally perturbs the received signal at the edge server and contributes to privacy protection.

In this paper, we consider the problem of FL with AirComp subject to client-level DP guarantee and aim to design a differentially private wireless FL scheme with high model accuracy and low communication and energy overhead. The first key challenge is the large noise magnitude required for protecting the privacy of each individual model update against attackers, which is proportional to the model dimension and results in a significantly degraded model accuracy. To address it, we propose to first compress the model update for reducing its model dimension and then use the compressed model update to design the transmit signal. As shown later, both model compression and the signal-superposition nature of wireless channel can reduce the noise magnitude required to provide a certain DP guarantee and achieve higher model accuracy. The second challenge is that when adding artificial noise to the local model updates for privacy protection, devices need to consume more energy for transmitting the noisy model updates to the server via AirComp as shown in~\cite{seif2020wireless}. If a strong DP guarantee is required, the additional energy consumption imposes a significant overhead on the energy-constrained edge device. Rather than adding artificial noise to the model updates when designing the transmit signal, we propose to adapt the transmit power and harness the channel noise in AirComp to naturally perturb the received signal at the server and provide intrinsic privacy. 

In summary, the contributions of this paper are as follows.
\begin{itemize}
    \item We propose a new wireless FL scheme to achieve DP with high model accuracy and low communication and energy overhead by harnessing the intrinsic channel noise, signal-superposition nature of wireless channel, and update compression in AirComp. 
    
    \item We analyze the convergence and privacy properties of the proposed scheme and show that the integrated design of model compression, signal superposition, and channel noise leads to strong DP protection while maintaining high model accuracy. 
    
    \item We conduct an extensive evaluation of the proposed scheme on benchmark FL datasets by comparing it with several baselines. The empirical results show that the proposed scheme can largely outperform the baselines in terms of model accuracy and energy efficiency given the same level of DP guarantee.  
\end{itemize}

%
\section{Related Work}

To improve the communication efficiency over wireless multiple-access channels in FL, AirComp-based FL has been proposed as a promising strategy by simultaneously transmitting model updates from all devices to the server \cite{zhu2019broadband,yang2020federated,amiri2020machine,amiri2020federated,zhang2021gradient}. Specifically, Zhu et al.~\cite{zhu2019broadband} considered an analog transmission strategy in FL, where each device is scheduled for transmission depending on the channel condition to improve the learning performance in the presence of fading channels. Yang et al.~\cite{yang2020federated} proposed a joint device selection and beamforming design while guaranteeing the mean squared error (MSE) to improve the learning task performance. Amiri et al.~\cite{amiri2020machine} proposed an algorithm in which devices pre-process the analog model updates through sparsification and quantization before transmissions. They further modify the analog communication scheme in \cite{amiri2020federated}, where the devices first sparsify their gradient estimates, and project the resultant error into a low-dimensional vector and transmit only the important gradient entries while accumulating the error from previous iterations. By taking gradient statistics into account, a gradient aware power control algorithm is developed in \cite{zhang2021gradient} to minimize the aggregation error. However, all of the aforementioned works did not consider the privacy issue when the central edge server is not trustworthy in AirComp-based FL. 

To mitigate the privacy risks for the model updates shared by the devices in AirComp-based FL, some recent studies \cite{seif2020wireless,mohamed2021privacy,SoneWFLLCDP,koda2020differentially,liu2020privacy} adopted AirComp-based FL with DP to prevent the privacy leakage by injecting random noise into the released model updates. In particular, the previous works~\cite{seif2020wireless,SoneWFLLCDP} utilized artificial noise to provide privacy guarantee when the inherent channel noise is not sufficient to satisfy the DP requirement. Accordingly, these strategies may lead to more energy consumption by transmitting artificial Gaussian noise. In~\cite{koda2020differentially}, an energy-efficient approach is proposed to scale down the transmit power instead of injecting artificial noise. Inspired by that, an adaptive power allocation scheme was proposed in~\cite{liu2020privacy} to achieve DP by utilizing intrinsic channel noise in both orthogonal multiple access (OMA) and nonorthogonal multiple access (NOMA) channels. Nevertheless, both works considered the full gradients transmission. Therefore, the communication cost is high due to the high dimension of DNNs. Meanwhile, to provide an unbiased estimation of model updates in AirComp-based FL, the updates need to be aligned by a coefficient related to the channel and power conditions before transmission~\cite{seif2020wireless,koda2020differentially,liu2020privacy}. Since both works \cite{koda2020differentially} and \cite{liu2020privacy} considered the full device participation, their power alignment coefficients are limited by the device with the worst channel and power condition. Furthermore, aforementioned works~\cite{seif2020wireless,mohamed2021privacy,SoneWFLLCDP,liu2020privacy} consider record-level DP, which protects the presence of a single data sample. As elaborated in \cite{McMahan2018learning,zhu2020voting}, it makes more sense to consider the notion of client-level DP in FL, which protects the presence of a device's entire dataset and is more challenging to achieve. Our work aims to achieve client-level DP while ensuring a high level of model accuracy in DNNs by jointly exploiting the opportunities of model compression and adaptive power control. 

\section{Preliminaries}
\begin{table}[t]
  \caption{Summary of main notations.}
  \label{tab:notations}
  \centering
  \begin{tabular}{cc}
    \toprule
    Notation & Definition\\
    \midrule
    $i, j$ &  Index for device\\
    $s$ & Index for local iteration\\
    $t$ & Index for global iteration\\
    $N$ & Total number of devices\\
    $[N]$ & \{1, 2, \ldots, $N$\}\\
    $\mathcal{S}^t$ & Set of selected devices in iteration $t$\\
    $\bm{\theta}_i^{t,s}$ &  Local model of device $i$\\ 
    ${D}_i$ & Local dataset of device $i$\\
    $f_i(\cdot)$ & Local objective function of device $i$ \\
    $\Delta_i^t$ & Model updates of device $i$\\
    $\mathbf{g}_i^{t,s}$ & Stochastic gradient of device $i$\\
    $\eta$ & Local learning rate\\
    $\tau$ & Aggregation period\\
    $C$ & Clipping threshold\\
    $\bm{h}_i^t$ & Channel gain between device $i$ and server\\ 
    $\mathbf{x}_{i}^{t}$ & Transmit signal of device $i$\\
    $\mathbf{z}^t$ & Random Gaussian noise\\
    $\mathbf{y}^t$ & Received signal by server\\
    $P_i$ & Transmission power limit of device $i$\\
    $\beta^t$ & Alignment coefficient\\
    $C_1$ & Bounded gradient coefficient\\
    $\zeta_i^{2}$ & Bounded variance coefficient\\
   $\gamma^2, \kappa^2$ & Bounded dissimilarity coefficients\\
   $\epsilon$, $\delta$ & differential privacy parameters\\
  \bottomrule
\end{tabular}
\end{table}
\subsection{Federated Learning}
\begin{figure}[t]
\centering
\includegraphics[width=0.48\textwidth]{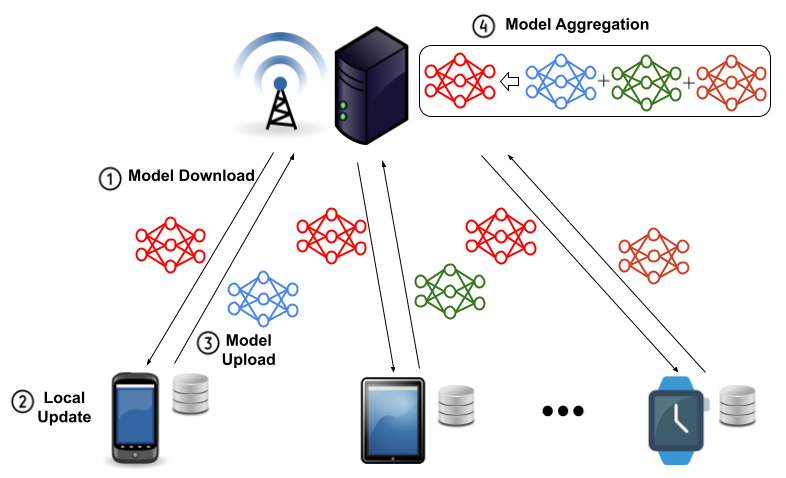}
\caption{FL system model.}\label{fig:system}
\end{figure}
As shown in Fig.~\ref{fig:system}, a typical FL system enables $N$ edge devices and a server to collaboratively solve an optimization problem of the form:
\begin{equation}\label{pro:fl}
\min_{\bm{\theta} \in \mathbb{R}^d} f(\bm{\theta}) := \frac{1}{N} \sum_{i=1}^N f_i(\bm{\theta}),
\end{equation}
where $f_i(\bm{\theta}) := (1/|D_{i}|)|\sum_{\xi \in D_{i}} l_i(\bm{\theta}; \xi)$ is the local objective function of device $i$, and $D_i$ is the local dataset of device $i$. Here $l_i$ is the (possibly non-convex) loss function defined by the learning task, and $\xi$ represents a data sample from $D_i$. The main notations used in the paper are summarized in Table~\ref{tab:notations}.

As a classic approach in FL optimization, Federated Averaging (FedAvg)~\cite{mcmahan2017communication} was proposed to solve~\eqref{pro:fl} by performing multiple steps of stochastic gradient descent (SGD) on each edge device and averaging the model updates periodically on the edge server. FedAvg can save communication rounds compared with distributed SGD and has been found to be simple and effective in many convex and non-convex settings empirically. More specifically, in the $t$-th FL round, FedAvg first sends the global model $\bm{\theta}^t$ to a subset of edge devices $\mathcal{S}^t$ with $r = |\mathcal{S}^t|$. Then each edge device $i \in \mathcal{S}^t$ computes a local model based on its local dataset by performing $\tau$ steps of SGD:
\begin{gather}
\bm{\theta}_{i}^{t,0} = \bm{\theta}^{t}, \label{eq_modl_init}\\
\bm{\theta}_{i}^{t, s} = \bm{\theta}_{i}^{t, s - 1} - \eta \mathbf{g}_i^{t,s-1}, \quad \forall s = 1, \ldots, \tau\label{eq_local_iter}
\end{gather}
where $\mathbf{g}_i^{t,s-1}$ is the stochastic gradient computed on a mini-batch $z_i$ randomly sampled from the local dataset $D_i$, and $\bm{\theta}_{i}^{t, s}$ is the local model of device $i$ at the $s$-th local iteration of round $t$. Next, devices upload their models $\bm{\theta}_{i}^{t, \tau}$ or model updates $\Delta_{i}^{t} := \bm{\theta}_{i}^{t, \tau} - \bm{\theta}^{t}$ to the server, which will then aggregate them to obtain a new global model:
\begin{equation}
\bm{\theta}^{t + 1} = \bm{\theta}^t + \frac{1}{r}\sum_{i \in \mathcal{S}^t} \Delta_i^t .
\end{equation}

\subsection{Differential Privacy}

DP provides a rigorous privacy concept to prevent the privacy leakage and has become the de-facto standard for measuring privacy risk \cite{dwork2014algorithmic}. Informally, DP mandates that the output distribution of an algorithm remains roughly the same when an individual's data is added or removed from the dataset. This ensures that an adversary with access to the statistics of the dataset cannot distinguish whether the target individual is present in the dataset or not. The formal $(\epsilon, \delta)$-DP is defined as follows.

\begin{definition}[$(\epsilon,\delta)$-DP\cite{dwork2014algorithmic}]\label{def:dp}
Given privacy parameters $\epsilon>0$ and $0\leq\delta<1$, a random mechanism $\mathcal{M}$ satisfies $(\epsilon,\delta)$-DP if for all neighboring datasets $D \simeq D^\prime$ and any subset of outputs $\mathcal{O} \subseteq range(\mathcal{M})$, we have
\begin{align}
\textup{Pr}[\mathcal{M}(D)\in \mathcal{O}]\leq e^{\epsilon}\textup{Pr}[\mathcal{M}(D^\prime)\in \mathcal{O}]+\delta.
\end{align}
When $\delta = 0$, we have $\epsilon$-DP, or Pure DP.
\end{definition}
The parameter $\epsilon$ is known as the \emph{privacy budget} that controls the trade-off between privacy and utility of the mechanism $\mathcal{M}$. A smaller $\epsilon$ provides a stronger privacy guarantee but typically results in a lower utility. The parameter $\delta$ is usually set to a small value to account for a probability that the upper bound $\epsilon$ fails. The details of definition of neighboring datasets $D \simeq D^\prime$ will be discussed in the next subsection.

The Gaussian mechanism is commonly used to achieve $(\epsilon,\delta)$-DP by injecting zero-mean Gaussian noise to the query output, the scale of which depends on the $\ell_2$-sensitivity of the query function. The definition of sensitivity is given as follows.

\begin{definition}[$\ell_2$-sensitivity\cite{dwork2014algorithmic}]
\label{def:sensitivity}
Let $h:\mathcal{D} \rightarrow \mathbb{R}^d$ be a query function over a dataset. The $\ell_2$-sensitivity of $h$ is defined as
\begin{equation}
    \psi(h)\coloneqq\max_{D \simeq D^\prime }\|h(D)-h(D^\prime)\|_2
\end{equation} where $D$ and $D^\prime$ are two neighboring datasets. 
\end{definition}

\begin{theorem}[Gaussian Mechanism \cite{dwork2014algorithmic}]\label{theo:gaussian_mechanism}
Let $h: \mathcal{D} \rightarrow \mathbb{R}^d$ be a query function with $\ell_2$-sensitivity $\psi(h)$. The Gaussian mechanism $\mathcal{M}(D) = h(D) + \mathcal{N}\left(0, \sigma^2\mathbf{I}_d\right)$ with $\sigma^2\geq{2 \ln(1.25/\delta)}\psi^2(h)/{\epsilon^2}$ satisfies $(\epsilon,\delta)$-DP for any $\delta \in (0, 1)$ and $\epsilon > 0$. 
\end{theorem}
In DP mechanisms, the privacy amplification property of DP~\cite{balle2018privacy} allows us to improve the privacy guarantees of DP algorithms without increasing the added noise. Specifically, by running a DP mechanism on a random subset of a dataset, it can provide stronger privacy than that of running on the entire dataset. The formal privacy amplification by subsampling is given as follows.
\begin{theorem}[Privacy Amplification by Subsampling\cite{balle2018privacy}]
\label{theo_priv_amply}
Suppose a mechanism $\mathcal{M}$ is $(\epsilon,\delta)$-DP over a dataset $D$ of size $m$. Consider the subsampling mechanism that given the set $D$ outputs a sample from the uniform distribution over all subsets $D_s\subseteq D$ of size $n$ $(n\leq m)$. Executing $\mathcal{M}$ on the subset $D_s$ guarantees $(\epsilon^{\prime}, {n}\delta/{m})$-DP, where $\epsilon^{\prime} = \log(1+{n}(e^{\epsilon}-1)/{m})$.
\end{theorem}

\subsection{Record-level and Client-level DP in FL}

Depending on how the neighboring datasets are defined, the DP definition can be applied to different granularities. Some prior works~\cite{abadi2016deep,liu2020fedsel,hu2021federated,hu2021concentrated,hu2020personalized} on differentially private FL deal with record-level DP that is defined as follows:
\begin{definition}[Record-level neighboring datasets]\label{def:record_level_neigh} 
Two datasets $D=\{D_{i}\}_{i\in[N]}$ and $D^\prime=\{D_{i}^\prime\}_{i\in[N]}$ are neighboring datasets if a device $j$'s dataset $D_{j}^\prime$ is constructed by adding or removing a single training data example $\xi$ from $D_j$. i.e., $D_{j}^\prime=D_{j}\cup\{\xi\}$ or $D_{j}^\prime=D_{j}\backslash\{\xi\}$, and $D_{i}^\prime=D_{i}$ for all $i\neq j$.
\end{definition}

Accordingly, only one training example's privacy is protected. Protecting individual examples is insufficient in many FL setting because one user may contribute many examples to the training dataset. The same example might be contributed multiple times by an individual user, but it should still be protected~\cite{McMahan2018learning}. In this paper, we consider the client-level DP to protect the privacy of entire examples of a client in the training dataset. The client-level neighboring datasets in FL can be defined as follows:

\begin{definition}[Client-level neighboring datasets]\label{def:client_level_neigh} 
Two datasets $D$ and $D^\prime$ are neighboring datasets if $D^\prime$ is constructed by adding or removing all of the examples associated with a single client $j$ from $D$, i.e., $D=\{D_{i}\}_{i\in[N]}$, $D^\prime=D\cup D_{j}$ or $D^\prime=D\backslash D_{j}$. This implies that $D$ and $D'$ only differ in one client's dataset.   
\end{definition}
Intuitively, client-level DP implies that a single client's contribution would not have a significant impact on the output distribution of the DP algorithm.

\begin{algorithm}[t]  
\textbf{Input:} Initial server model $\bm{\theta}^0$, aggregation period $\tau$, total rounds $T$, sample size $r$, clipping threshold $C$, noise magnitude $\sigma$, and learning rate $\eta$.\\
\textbf{Output:} Final global model $\bm{\theta}^{T}$ 
\caption{DP-FedAvg~\cite{McMahan2018learning} }\label{algorithm-dpfedavg}
\begin{algorithmic}[1]
    \FOR{ $t = 0, \ldots, T - 1$}
    \STATE Uniformly sample a set $\mathcal{S}^t\subseteq[N]$ with $r=|\mathcal{S}^t|$\label{server_sample}
    \STATE Broadcast $\bm{\theta}^t$ to all clients in $\mathcal{S}^t$\label{server_broad}
    \FOR{each device $i \in \mathcal{S}^t$ \textbf{in parallel}}
        \STATE $\bm{\theta}_i^{t,0}\gets$ $\bm{\theta}^t$\label{device_init}
        \FOR{$s=1,\dots,\tau$}\label{device_sgd_start}
            \STATE 
            Compute a stochastic gradient $\mathbf{g}_{i}^{t,s-1}$ over a mini-batch $\xi_i$ sampled from $D_i$
            \STATE $\bm{\theta}_{i}^{t,s}\gets$ $\bm{\theta}_{i}^{t,s-1}-\eta\mathbf{g}_{i}^{t,s-1}$
        \ENDFOR\label{device_sgd_end}
        \STATE $\Delta_{i}^{t}\gets \bm{\theta}_i^{t,\tau} - \bm{\theta}^{t}$\label{device_cal_upd}
        \STATE $\tilde{\Delta}_{i}^{t} \gets \Delta_{i}^{t}/\max\left(1,\frac{\norm{\Delta_{i}^{t}}_2}{C}\right)+\mathcal{N}(0,C^2\sigma^2\mathbf{I}_d/r)$\label{device_clip_noise}
    \ENDFOR
    \STATE $\bm{\theta}^{t + 1} \gets \bm{\theta}^t + \frac{1}{r}\sum_{i\in\mathcal{S}^t}\tilde{\Delta}_{i}^{t}$\label{server_aggr}
\ENDFOR
\end{algorithmic}
\end{algorithm}

\subsection{DP-FedAvg: Achieving Client-level DP in FL}
To provide client-level DP in FL under a ``honest-but-curious'' edge server, DP-FedAvg\cite{McMahan2018learning} can be adapted to this setting by perturbing the model updates locally before uploading them to the server. Specifically, as shown in Algorithm~\ref{algorithm-dpfedavg}, DP-FedAvg consists of the following steps in each FL round $t$:
\begin{enumerate}
    \item Server sends the global model to a randomly sampled subset of devices (lines~\ref{server_sample}-\ref{server_broad});
    \item Each device initializes its local model to be the received global model (line~\ref{device_init}), performs $\tau$ steps of SGD  (lines~\ref{device_sgd_start}-\ref{device_sgd_end}) and computes its local model update (line~\ref{device_cal_upd});
    \item Each device clips the norm of model updates $\Delta_{i}^{t}$ by a threshold $C$ and adds Gaussian noise to its bounded local model update (line~\ref{device_clip_noise}); 
    \item Server aggregates the perturbed local model updates received from edge devices to update the global model (line~\ref{server_aggr}). 
\end{enumerate}
    
Although DP-FedAvg ensures client-level DP, it suffers from large model accuracy degradation and high energy consumption of edge devices. In this paper, we propose to integrate compression and intrinsic channel noise to achieve client-level DP while achieving high communication and energy efficiency and model accuracy in wireless FL.

\section{Model and Privacy Goal}

In this subsection, we first introduce the wireless communication model in AirComp-based FL and then describe the threat model and privacy goal.

\subsection{Wireless Communication Model}

\begin{figure}[t]
\centering
\includegraphics[width=0.48\textwidth]{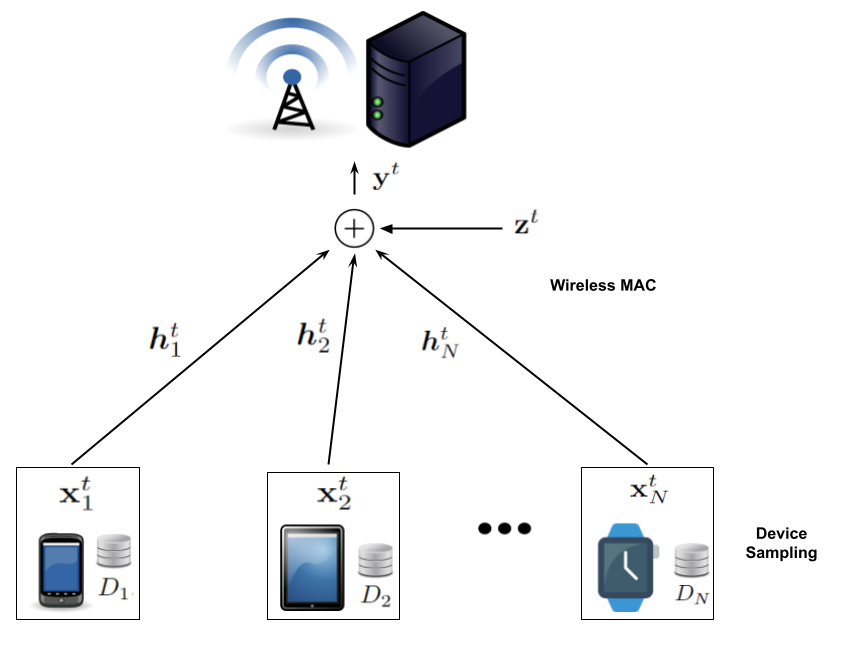}
\caption{Wireless communication model in AirComp.}\label{fig:wfl_system}
\end{figure}

As shown in Fig.~\ref{fig:wfl_system}, the edge server communicates with all devices via a wireless flat-fading channel. To simplify the analysis, we assume the download link is ideal so that the download signal is not distorted since the edge server has a much stronger transmission power compared to the devices. 
The channel gain between device $i$ and the edge server at round $t$ is denoted as a complex value $\bm{h}_i^t=|h_i^t|e^{j\phi_i^t}$. In this paper, we assume the channel state information (CSI) to be constant within each FL round, but may vary over rounds. We also assume that the CSI is perfectly known at the server and edge devices as done in the literature \cite{koda2020differentially} and leave the case of imperfect CSI for future work.

We consider that the channel is split into $K$ subcarriers under OFDMA, and each device can simultaneously transmit at most $K$ elements of local signal $\mathbf{x}_{i}^{t}$ using the subcarriers to the server with analog transmissions in one time slot. For instance, in 5G systems, a 50 MHz channel with 15 KHz carrier spacing provides 3300 subcarriers during 10 ms \cite{elgabli2021harnessing}. Provided that we have accurate channel-gain precoding for phase compensation and strict synchronization among the participating devices \cite{shao2021federated}, in the model upload stage of the $t$-th round, each device $i$ first designs the transmitted signal as $e^{-j\phi_i^t}\mathbf{x}_{i}^{t}$, where $e^{-j\phi_i^t}$ is the local phase correction performed by the device $i$. Then, each coordinate of the local model update is assigned to a specific subcarrier and then transmitted via a wireless MAC simultaneously. By exploiting the signal-superposition nature of wireless channel, the edge server then receives the coordinates transmitted by different devices over the same set of subcarriers in the form of an aggregated sum. Specifically, the received signal $\mathbf{y}^t$ over $K$ subcarriers at the edge server in the $t$-th FL round is
\begin{align}\label{eq_rece_sign}
\mathbf{y}^t & = \sum_{i \in \mathcal{S}^t} \bm{h}_i^te^{-j\phi_i^t}\mathbf{x}_{i}^{t} + \mathbf{z}^t\notag\\
& = \sum_{i \in \mathcal{S}^t} |h_{i}^t|\mathbf{x}_{i}^{t} + \mathbf{z}^t,
\end{align}
where $\mathbf{z}^t\sim\mathcal{N}(0,\sigma_0^2\mathbf{I}_K) $ is the random Gaussian noise with variance $\sigma_0^2$ over $K$ subcarriers.

Since edge devices participating in FL are resource-constrained, we have a transmission power limit for each edge device $i$:
\begin{equation}\label{eq_powe_cont}
\text{(Power Limit)}  \quad\mathbb{E} \norm{\mathbf{x}_i^t}_2^2 \leq P_i,\quad \forall t.
\end{equation}

\subsection{Adversarial Model and Privacy Goal} 

In this paper, we consider the edge server to be ``honest-but-curious''. That is, the edge server is curious about a specific device's local dataset and intends to infer information from the shared messages, but honestly follows the protocols involving the training process. Moreover, there may exist a third party that can observe the global model broadcasted by the server in each round.
The privacy goal of this paper is to ensure that the edge server or third party cannot infer much about a device's local dataset by observing the received global model update in each round.

\section{PFELS: Private Federated Edge Learning with Sparsification}

In this section, we propose a new wireless FL scheme named Private Federated Edge Learning with Sparsification (PFELS) to achieve client-level DP while keeping high model accuracy and improving communication and energy efficiency at wireless edge. We also formulate the optimal power control design in PFELS as a constrained optimization problem.

\subsection{Proposed Learning Scheme}

To reduce communication cost and energy consumption, the key idea of PFELS is that each device sparsifies the model update first before transmitting it to the server via wireless MAC. Moreover, rather than adding artificial noise to the local model updates for privacy protection, which will consume extra transmission energy, PFELS relies on the inherent channel noise and adaptive power scaling to achieve DP in FL.

\begin{algorithm}[t]
\textbf{Input:} Initial server model $\bm{\theta}^0$, local update period $\tau$, number of sampled clients $r$, number of total rounds $T$, and learning rate $\eta$.\\
\textbf{Output:} Final global model $\bm{\theta}^{T}$ 
\caption{PFELS Algorithm.}\label{algorithm-2}
\begin{algorithmic}[1]
    \FOR{ $t = 0, \ldots, T - 1$}
    \STATE Uniformly sample a set $\mathcal{S}^t\subseteq[N]$ with $r=|\mathcal{S}^t|$ without replacement\label{alg_lin_samp}
    \STATE Generate a $\randk_k$ projection matrix $\mathbf{A}^{t} \in \{0, 1\}^{k\times d}$\label{alg_lin_randk}
    \STATE Broadcast $\bm{\theta}^t$ and $\mathbf{A}^{t}$ to all devices in $\mathcal{S}^t$ \label{alg_lin_set}
    \FOR{each device $i \in \mathcal{S}^t$ \textbf{in parallel}}
        \STATE $\bm{\theta}_i^{t,0}\gets$ $\bm{\theta}^t$\label{alg_lin_loc_ini}
        \FOR{$s=1,\dots,\tau$}
            \STATE Compute a stochastic gradient $\mathbf{g}_{i}^{t,s-1}$ over a mini-batch $\xi_i$ sampled from $D_i$
            \STATE $\bm{\theta}_{i}^{t,s}\gets$ $\bm{\theta}_{i}^{t,s-1}-\eta\mathbf{g}_{i}^{t,s-1}$  \label{alg_lin_loc_end}
        \ENDFOR 
        \STATE $\Delta_{i}^{t} \gets \bm{\theta}_i^{t,\tau} - \bm{\theta}^{t}$\label{alg_lin_loc_upd}
        \STATE $\mathbf{x}_{i}^t \gets \alpha_{i}^t \mathbf{A}^{t}\Delta_{i}^t$\label{alg_lin_sns}
    \ENDFOR
    \STATE $\mathbf{y}^t \gets \sum_{i \in \mathcal{S}^t}|h_i^t|\mathbf{x}_i^t + \mathbf{z}^t$\label{alg_lin_yt}
    \STATE $\hat{\Delta}^t \gets \frac{(\mathbf{A}^{t})^{\intercal}}{r\beta^t}\mathbf{y}^{t}$\label{alg_lin_hatDel}
    \STATE $\bm{\theta}^{t + 1} \gets \bm{\theta}^t + \hat{\Delta}^t$ \label{alg_lin_global}
\ENDFOR
\end{algorithmic}
\end{algorithm}

The pseudo-code for the proposed PFELS is provided in Algorithm~\ref{algorithm-2}. At each round $t$, the server randomly samples a set $\mathcal{S}^t$ of $r$ devices with uniform probability at random without replacement (line~\ref{alg_lin_samp}). After that, a $\randk_k$ projection matrix $\mathbf{A}^{t}\in \{0, 1\}^{k\times d}$ is generated by the server according to \eqref{def_rand_matx} where $k<d$ (line~\ref{alg_lin_randk}). Note that the error compensation technique~\cite{karimireddy2019error,lin2018deep,tang2019doublesqueeze} has been proposed to accelerate model convergence under compression and can be integrated into PFELS to further improve accuracy. Here, $\mathbf{A}^{t}$ is utilized to project a $d$-dimensional vector into a $k$-dimensional vector such that $\mathbf{A}^t \mathbf{V}\in\mathbb{R}^k$ for any $ \mathbf{V}\in\mathbb{R}^d$. In particular, let $\Omega_k = \binom{[d]}{k}$ denote the set of all $k$-element subsets of $[d]$, and a subset $\omega=\{\omega_{1},\dots,\omega_{k}\}$ is chosen uniformly and randomly from $\Omega_k$. Then, the $\randk_k$ projection matrix $\mathbf{A}^t \in\mathbb{R}^{k\times d}$ is defined as follows:
\begin{align}\label{def_rand_matx}
& [\mathbf{A}^t]_{m,n}:=
\begin{cases} 
1, & \text{if } n = \omega_{m}\\ 
0, & \text{otherwise.}
\end{cases}
\end{align}
Then, the global model $\bm{\theta}^t$ and projection matrix $\mathbf{A}^t$ are shared with all selected clients (line~\ref{alg_lin_set}). In practice, to avoid an extra communication cost of $\mathbf{A}^t$, the coordinate set $\omega$ can be selected from $\Omega_k$ distributively by the server and clients via pseudo-random generators with the same seed.

After receiving the global model $\bm{\theta}^t$ and random projection matrix $\mathbf{A}^{t}$ from the server, each device $i\in\mathcal{S}^t$ initializes its local model to $\bm{\theta}^t$ and performs $\tau$ step of SGD to update its local model in parallel (lines~\ref{alg_lin_loc_ini}-\ref{alg_lin_loc_end}). Then, each device $i$ calculates its local model update $\Delta_{i}^{t}$ (lines~\ref{alg_lin_loc_upd}) and designs the transmit signal $\mathbf{x}_{i}^t\in\mathbb{R}^k$ by sparsifying its local model update using $\mathbf{A}^{t}$ and then scaling it (lines~\ref{alg_lin_sns}) as
\begin{equation}\label{eq:spar_signal}
\mathbf{x}_{i}^t = \alpha_{i}^t \mathbf{A}^{t}\Delta_{i}^t,
\end{equation}
where $\alpha_{i}^t\in\mathbb{R}$ is the power scaling factor to be optimized later. 

After that, all devices in $\mathcal{S}^t$ send their transmit signals to the server via a channel with additive Gaussian noise $\mathbf{z}^t$. Next, according to the channel model~\eqref{eq_rece_sign} and input signals~\eqref{eq:spar_signal}, the signal received at the edge server becomes
\begin{align}\label{eq:receive_signal}
\mathbf{y}^{t} =& \sum_{i \in \mathcal{S}^t} |h_{i}^t|\mathbf{x}_{i}^{t} + \mathbf{z}^t,
\end{align}
where $\mathbf{z}^t\sim\mathcal{N}(0, \sigma_0^2 \mathbf{I}_k)$ is the additive Gaussian channel noise at the edge server (line~\ref{alg_lin_yt}). In order to align transmitted local model updates, we have the following power alignment constraints:
\begin{align}\label{eq_pow_align}
&\text{(Power Alignment)} & |h_{i}^t|\alpha_{i}^t = \beta^t, \quad \forall i, t,
\end{align}
where the alignment coefficient $\beta^t$ is to be properly designed later in Section~\ref{sec_opt}. Finally, the server can estimate a $d$-dimensional aggregated model update $\hat{\Delta}^t$ from the received $k$-dimensional signal $\mathbf{y}^t$ (line~\ref{alg_lin_hatDel}) as
\begin{align}\label{eq:receive_signal_2}
\hat{\Delta}^t=\frac{(\mathbf{A}^{t})^{\intercal}}{r\beta^t}\mathbf{y}^{t}
= \frac{1}{r}\sum_{i\in\mathcal{S}^t} (\mathbf{A}^{t})^{\intercal}\mathbf{A}^{t}\Delta_{i }^t + \frac{(\mathbf{A}^{t})^{\intercal}\mathbf{z}^{t}}{r\beta^t},
\end{align}
where $(\mathbf{A}^{t})^{\intercal}\mathbf{A}^{t}\Delta_{i }^t\in \mathbb{R}^{d}$ is the sparsified local model update by only keeping the $k$ coordinates of the orignal model update for device $i$. Then, the edge server uses the estimated aggregated model update to compute its global model for the next round (line~\ref{alg_lin_global}). 

We show that an unbiased estimate of the true aggregated model update$\sum_{i\in\mathcal{S}^t}\Delta_{i }^t/r$ in each round $t$ can be recovered via $\hat{\Delta}^t$ as follows:
\begin{lemma}\label{lemma_estimatation}
Given a parameter $k\in[d]$ and a $\randk_k$ projection matrix $\mathbf{A}^t\in\mathcal{R}^{k\times d}$ generated from the active subset $\omega$, it holds that
\begin{gather}
    \mathbb{E}_{\omega}[\hat{\Delta}^t]=\frac{k}{d}\sum_{i\in\mathcal{S}^t}\frac{\Delta_{i }^t}{r}.   
\end{gather}
\end{lemma}
\begin{IEEEproof}
The proof is provided in Appendix~\ref{append_proof_lemma_estimatationp} in the
supplementary.
\end{IEEEproof}

\subsection{Optimal Power Control Design in PFELS}
The design goal of power control in PFELS is to minimize the training loss while satisfying privacy and energy constraints with adaptive power control $\{\beta^t\}_{t\in[T-1]}$. We observe that the effective noise added to the model update depends on $\{\beta^t\}_{t\in[T-1]}$ according to \eqref{eq:receive_signal_2} and will determine the loss $f(\bm{\theta}^T)$ after $T$ rounds. Therefore, our goal can be represented as solving the following optimization problem:

\begin{equation}
    \mathbf{P1}\quad \min_{\{\beta^t\}_{t\in[T-1]}} \mathbb{E} [f(\bm{\theta}^T)] 
\end{equation}
while satisfying privacy constraint, power limit constraint~\eqref{eq_powe_cont}, and power alignment constraint~\eqref{eq_pow_align}.

Solving the aforementioned problem $\mathbf{P1}$ faces several challenges. First, we need to find out how the control decisions $\{\beta^t\}_{t\in[T-1]}$ affect the loss $f(\bm{\theta}^T)$. Generally, there is no explicit mathematical expression to capture the relationship between $\{\beta^t\}_{t\in[T-1]}$ and $f(\bm{\theta}^T)$. To address it, we propose to analyze the convergence properties of PFELS in the next section and use the convergence error bound to substitute the loss function as the surrogate objective in the optimization problem. Second, we need to carefully design $\{\beta^t\}_{t\in[T-1]}$ to guarantee a certain level of client-level DP. Therefore, we provide the privacy analysis in the next section and model the privacy constraint as a function of $\beta^t$.

\section{Privacy and Convergence Analysis}

In this section, we provide the client-level DP guarantee and convergence properties of PFELS under the general non-convex and non-IID setting. Before stating our theoretical results, we make the following assumptions:
\begin{assumption}[Bounded Gradient]\label{ass_Bound_Grad} 
The stochastic gradient in Algorithm~\ref{algorithm-2} for any client is bounded by a constant $C_1 > 0$, i.e., $\norm{\mathbf{g}_{i}^{t,s}}_{2}\leq C_1$ for any $i$, $t$, and $s$.
\end{assumption}
\begin{assumption}[Smoothness]
\label{ass:smoothness}
Each local objective function $f_i:\mathbb{R}^d\rightarrow\mathbb{R}$ is $L$-smooth for all $i\in [N]$, i.e.,
\begin{equation}
    \norm{\nabla f_i(\bm{\theta}) - \nabla f_i(\bm{\theta}^\prime)}_2^2 \leq L \norm{\bm{\theta} - \bm{\theta}^{\prime}}_2^2,    \; \forall \bm{\theta},\bm{\theta}^{\prime}\in \mathbb{R}^d. 
\end{equation}
\end{assumption}

\begin{assumption}[Unbiased Gradient and Bounded Variance]\label{ass:gradient}
The local mini-batch stochastic gradient is an unbiased estimator of the local gradient: $\mathbb{E}[\mathbf{g}_i(\bm{\theta})] = \nabla f_i(\bm{\theta})$ and has bounded variance: $\mathbb{E}[\norm{ \mathbf{g}_i(\bm{\theta}) - \nabla f_i(\bm{\theta})}_2^2] \leq \zeta_i^2, \forall \bm{\theta} \in \mathbb{R}^d, i \in [N]$, where the expectation is over all the local mini-batches. We also denote $\bar{\zeta}^2\coloneqq (1/N)\sum_{i=1}^N \zeta_i^2$ for convenience.
\end{assumption}

\begin{assumption}[Lower Bounded]\label{ass:lowerbounded}
There exists a constant $f_{\inf}$ such that
\[
f(\bm{\theta}) \geq f_{\inf},\quad \forall \bm{\theta}\in\mathbb{R}^d.
\]
\end{assumption}

\begin{assumption}[Bounded Dissimilarity]\label{ass:bound_dissimi}
There exist constants $\gamma^2\geq  1, \kappa^2 \geq 0$ such that $(1/N)\sum_{i=1}^{N}\norm{\nabla f_i(\bm{\theta})}_2^2\leq \gamma^2\norm{(1/N)\sum_{i=1}^{N}\nabla f_i(\bm{\theta})}_2^2+\kappa^2$. If the data across all devices are IID, then $\gamma^2=1$ and $\kappa^2=0$.
\end{assumption}

Note that Assumption~\ref{ass_Bound_Grad} has been commonly made in differentially private ML literature \cite{geyer2017differentially,wei2020federated,McMahan2018learning,abadi2016deep}, which can be ensured by the gradient clipping \cite{abadi2016deep}. Assumptions \ref{ass:smoothness}--\ref{ass:lowerbounded} are standard in the analysis of SGD \cite{bottou2018optimization}. Assumption~\ref{ass:bound_dissimi} is commonly used in the FL literature \cite{koloskova2020unified, wang2020tackling} to capture the dissimilarities of local objectives due to data heterogeneity.

\subsection{Privacy Analysis}

In this subsection, we provide the client-level DP analysis of PFELS.
Theorem~\ref{theo:gaussian_mechanism} indicates that the DP guarantee $(\epsilon, \delta)$ depends on the sensitivity of the query function on the private dataset. In PFELS, according to~\eqref{eq:spar_signal},~\eqref{eq:receive_signal} and~\eqref{eq_pow_align}, the received signal at the edge server can be rewritten as: 
\begin{align}
    \mathbf{y}^{t} = \sum_{i \in \mathcal{S}^t} \beta^t\mathbf{A}^t\Delta_{i}^t + \mathbf{z}^t.
\end{align}
Thus, the edge server only knows the sum of local model updates $\sum_{i\in\mathcal{S}^t}\beta^t\mathbf{A}^t\Delta_{i}^t$, which depends on the private dataset. Therefore, we need to analyze the sensitivity of the sum of local model updates.
Assume the client-level neighboring datasets $D=\{D_i\}_{i\in\mathcal{S}^t}$ and $D^{\prime}=D\cup D_{e}$ or $D^{\prime}=D\backslash D_{e}$ that differ in client $e$'s dataset at round $t$, and the corresponding device datasets are $\mathcal{S}^t$ and $(\mathcal{S}^t)^\prime$, respectively. According to Definition~\ref{def:sensitivity}, we have the following lemma: 
\begin{lemma}[Sensitivity in PFELS]
\label{lemma_sens}
Under Assumption~\ref{ass_Bound_Grad}, the $\ell_2$-norm sensitivity of the sum of local model updates is
\begin{equation*}\label{eq:sensitivity_pfels}
    \psi_{\Delta}=\max_{D \simeq D^\prime}\|\sum_{i\in\mathcal{S}^t } \beta^{t}\mathbf{A}^t\Delta_{i }^t  - \sum_{i^{\prime}\in(\mathcal{S}^t)^{\prime} } \beta^{t}\mathbf{A}^t\Delta_{i^{\prime} }^t\|_2 \leq \beta^t\eta\tau C_1.
\end{equation*}
\end{lemma}
\begin{IEEEproof}
According to the local model updating rule~\eqref{eq_modl_init} and~\eqref{eq_local_iter}, each device $i$ initializes its model with $\bm{\theta}^t$ and performs $\tau$ local SGD to compute the final local model $\bm{\theta}_{i}^{t,\tau}$. Therefore, we have:
\begin{equation}\label{eq_round_sgd}
    \bm{\theta}_{i}^{t,\tau} = \bm{\theta}^{t}-\eta\sum_{s=1}^{\tau}\mathbf{g}_{i}^{t,s-1}. 
\end{equation}
Then, for the difference of the received aggregate by the server between neighboring sets $\mathcal{S}^t$ and $(\mathcal{S}^t)^{\prime}$ that differ in one client index $e$, its $\ell_2$-norm is:
\begin{subequations}
\begin{align}
  \big\| \sum_{i\in\mathcal{S}^t } \beta^{t}\mathbf{A}^t\Delta_{i }^t - &
    \sum_{i^{\prime}\in(\mathcal{S}^t)^{\prime} } \beta^{t}\mathbf{A}^t\Delta_{i^{\prime} }^t\big\|_2  = \norm{\beta^{t}\mathbf{A}^t\Delta_{e}^{t}}_2\\
    & = \beta^t\norm{\mathbf{A}^t(\bm{\theta}_{e}^{t, \tau} - \bm{\theta}^{t})}_2\\
    & = \beta^t\eta\|\mathbf{A}^t\sum_{s=1}^{\tau} \mathbf{g}_{e}^{t, s - 1}\|_2\label{eq_upd_to_grad}\\
    &\leq \beta^t\eta\sum_{s=1}^{\tau}\norm{\mathbf{A}^t\mathbf{g}_{e}^{t, s - 1}}_2
\end{align}
\end{subequations}
where~\eqref{eq_upd_to_grad} follows from~\eqref{eq_round_sgd}. $\mathbf{A}^t\mathbf{g}_{e}^{t, s - 1}$ means randomly keeping $k$ coordinates of $\mathbf{g}_{e}^{t, s - 1}$, and thus its upper bound depends on the upper bound of $\mathbf{g}_{e}^{t, s - 1}$. Thus, from Assumption~\ref{ass_Bound_Grad}, we have the final result.
\end{IEEEproof}

Next, we analyze the client-level DP guarantee of PFELS. Informally, according to Theorem~\ref{theo:gaussian_mechanism}, the Gaussian mechanism can provide $(\epsilon,\delta)$-DP. Based on that, considering the local model updates are performed on a random set of devices in PFELS, we can employ Theorem~\ref{theo_priv_amply} to provide a tighter $(\epsilon,\delta)$-DP bound for PFELS. Formally, we have the following theorem:

\begin{theorem}
\label{theo_dp_perRound}
Suppose that the client set $\mathcal{S}^t$ of size $r$ is uniformly sampled at random without replacement from $[N]$. If the power alignment parameter $\beta^t$ satisfies:
\begin{align}\label{eq_noise_low_bound}
C_2\beta^t\leq \epsilon,
\end{align}
where 
\begin{equation}
C_2 = \frac{2\sqrt{2}\eta\tau C_1r\sqrt{\log\left(1.25r/(N\delta)\right)}}{N\sigma_0},
\end{equation}
then each round of the PFELS algorithm guarantees $(\epsilon,\delta)$-DP for any $\epsilon\in(0,1)$ and $\delta\in(0,1)$.
\end{theorem}
\begin{IEEEproof}
In each round of PFELS, the intrinsic channel noise $ \mathbf{z}^t$ follows the distribution of $\mathcal{N}(0, \sigma_0^2 \mathbf{I}_k)$, and $r$ clients are uniformly sampled at random without replacement from $[N]$. According to Theorem~\ref{theo:gaussian_mechanism} and Theorem~\ref{theo_priv_amply}, the sensitivity $\psi_{\Delta}^2 \leq \sigma_0^2\epsilon_0^2/(2\log{(1.25/\delta_0)})$ can guarantee $(\log(1+(e^{\epsilon_0}-1)r/N), \delta_0 r/N)$-DP. For $\epsilon_0\in(0, 1)$, we have the following inequality:
\begin{equation}\label{in_privacy}
\log(1+\frac{r}{N}(e^{\epsilon_0}-1)) < \frac{r}{N}(e^{\epsilon_0} - 1) < \frac{2r}{N}\epsilon_0,
\end{equation}
where~\eqref{in_privacy} follows from the facts that $\log(1+x)<x$ for any $x>0$ and $e^x-1<2x$ for any $x\in(0,1)$.
Therefore, the sensitivity $\psi_{\Delta}^2 \leq \sigma_0^2\epsilon_0^2/(2\log(1.25/\delta_0))$ guarantees $(2r\epsilon_0/N, \delta_0{r}/{N})$-DP. In other words, to guarantee $(\epsilon, \delta)$-DP where $\epsilon = 2r\epsilon_0/N$ and $\delta = \delta_0{r}/{N}$, the sensitivity need to satisfy
\begin{equation*}
\psi_{\Delta}^2\leq\frac{\sigma_0^2N^2\epsilon^2}{8r^{2}\log{(1.25r/(N\delta))}}.
\end{equation*}
By replacing the $\ell_2$-norm sensitivity $\psi_{\Delta}$ by Lemma~\ref{lemma_sens}, we have:
\begin{equation*}
(\beta^t)^2\eta^2\tau^2 C_1^2\leq\frac{\sigma_0^2N^2\epsilon^2}{8r^{2}\log{(1.25r/(N\delta))}}.
\end{equation*}
Let $C_2$ denote $2\sqrt{2}\eta\tau C_1r\sqrt{\log\left(1.25r/(N\delta)\right)}/{(N\sigma_0)}$, we arrive at the conclusion.  
\end{IEEEproof}

\subsection{Convergence Analysis}
In this subsection, we provide the convergence analysis of PFELS under the non-convex and non-iid setting in Theorem~\ref{theorem_conv}. We only provide the proof sketches here and put the detailed proofs to the supplementary. 

\begin{lemma}[Iterate Decomposition]\label{lemma:iterate_t1_t2} Under Assumption~\ref{ass:smoothness},  the iterate of Algorithm~\ref{algorithm-2} satisfies:
    \begin{align*}
        \mathbb{E}_t  &[f(\bm{\theta}^{t+1}-f(\bm{\theta}^t)] \leq  L\eta^2\tau^2(\lambda_k+1)(\bar{\zeta}^2+2\kappa^2)+ \frac{\eta\tau\lambda_k\bar{\zeta}^2}{2\pi} \\
        &+ \left( \frac{\eta L^2}{2} +{2L^3\eta^2\tau(\lambda_k+1)}\right)\frac{1}{N}\sum_{i=1}^{N}\sum_{s=1}^{\tau}\mathbb{E}_t\norm{\bm{\theta}^t-\bm{\theta}_{i}^{t,s-1}}^2\\
         &+\left( -\frac{\eta\tau (1-\pi)}{2}+2L\eta^2\tau^2\gamma^2(\lambda_k+1) \right)\mathbb{E}_t\norm{\nabla f(\bm{\theta}^t)}^2\\
         &+ \frac{Lk\sigma_0^2}{2r^3(\beta^t)^2},
    \end{align*}
where $\lambda_k:=1-k/d $, $\pi>0$ is a constant.
\end{lemma}

\begin{IEEEproof}
    The proof is provided in Appendix~\ref{append_proof_iteate_decomp} in the supplementary.
\end{IEEEproof}
Lemma~\ref{lemma:iterate_t1_t2} provides the composition of per-round loss drift bound. The local parameter divergence $\|\bm{\theta}^t-\bm{\theta}_{i}^{t,s-1}\|^2$, $i\in[N]$ provides a hint on how to derive the convergence property of PFELS. Next, we will provide an upper bound on the local parameter divergence.

\begin{lemma}[Bounded Local Divergence]
\label{lemma_boud_locl_divg} 
Under Assumptions \ref{ass:smoothness}, \ref{ass:gradient} and \ref{ass:bound_dissimi},
the local model difference at round $t$ is bounded as follows:
\begin{multline*}
    \frac{1}{N}\sum_{i=1}^{N}\mathbb{E}_{t}\norm{\bm{\theta}^t-\bm{\theta}_{i}^{t,s-1}}^2 \leq16\eta_{l}^2\tau^2\gamma^2\mathbb{E}_t\norm{\nabla f(\bm{\theta}^t)}^2\\
    + 16\eta^2\tau^2\kappa^2+4\tau\eta^2\bar{\zeta}^2.
\end{multline*}
\end{lemma}
\begin{IEEEproof}
    The proof is provided in Appendix~\ref{append_proof_bound_locl_divg} in the supplementary.
\end{IEEEproof}
Lemma~\ref{lemma_boud_locl_divg} shows the upper bound of local divergence error. Combining Lemmas~\ref{lemma:iterate_t1_t2},~\ref{lemma_boud_locl_divg} and choosing a proper learning rate, we have the following convergence bound:
\begin{theorem}[Convergence of PFELS] \label{theorem_conv}
Under Assumptions \ref{ass:smoothness}-\ref{ass:bound_dissimi}, if the local learning rate satisfies $\eta\leq\min\{1/(24\tau L(\lambda_k+1)\gamma^2),1/(4\tau L\sqrt{4\gamma^2+2}),1/12\tau L\}$, then the iterations of PFELS after $T$ rounds satisfy:
\begin{align}\label{in_theo_conv}
    \frac{1}{T}\sum_{t=0}^{T-1}\mathbb{E}_t\norm{\nabla f(\bm{\theta}^t)}^2\leq&  \frac{8(f(\bm{\theta}^{0})-f_{\inf})}{T\eta\tau}+8\eta\tau L(3\kappa^2+2\bar{\zeta}^2)\notag\\
    &+8\left(\eta\tau L(2\kappa^2+\bar{\zeta}^2)+\frac{3\bar{\zeta}^2}{2}\right)\frac{\lambda_k}{r}\notag\\
    &+\frac{4Lk\sigma_0^2}{\eta\tau r^3T}\sum_{t=0}^{T-1}\frac{1}{(\beta^t)^2},
\end{align}
where $\lambda_k:=1-k/d $.
\end{theorem}
\begin{IEEEproof}
When $\pi < 1/2$ and 
\begin{equation}
    \eta\leq\min\{ \frac{1}{4\tau L\sqrt{4\gamma^2+2} }, \frac{1}{12\tau L}, \frac{1-2\pi}{8\tau L (\lambda_k+1)\gamma^2}\},\notag
\end{equation} 
we have:
\begin{gather}
    \left( -\frac{\eta\tau (1-\pi)}{2}+2L\eta^2\tau^2\gamma^2(\lambda_k+1) \right) \leq -\frac{\eta\tau}{4}\label{in_conv_eta1}\\
    2L^3\eta^2\tau(\lambda_k+1)\leq\frac{\eta L^2}{4\gamma^2}\label{in_conv_eta2}\\ 4\eta^3\tau^3L^2(2\gamma^2+1)\leq\frac{\eta\tau}{8}, 12\eta\tau L\leq 1.\label{in_conv_eta3}
\end{gather}
Substituting Lemma~\ref{lemma_boud_locl_divg} into Lemma~\ref{lemma:iterate_t1_t2}, we have:
\begin{align}
    \mathbb{E}_t&[f(\bm{\theta}^{t+1}) -f(\bm{\theta}^t)]  \leq {L\eta^2\tau^2(\lambda_k+1)}(2\kappa^2+\bar{\zeta}^2)+ \frac{\eta\tau\lambda_k\bar{\zeta}^2}{2\pi} \notag\\
    &\quad+ \left( \frac{\eta L^2}{2} +{2L^3\eta^2\tau(\lambda_k+1)}\right)\sum_{s=1}^{\tau} [16\eta^2\tau^2\gamma^2\left\| \nabla f(\bm{\theta}^t)\right\|^2 \notag\\
    &\quad+16\eta^2\tau^2\kappa^2+ {4\tau\eta^2}\bar{\zeta}^2]+ \frac{Lk\sigma_0^2}{2r^3(\beta^t)^2}\notag\\
    &\quad+\left( -\frac{\eta\tau (1-\pi)}{2}+2L\eta^2\tau^2\gamma^2(\lambda_k+1) \right)\mathbb{E}_t\norm{\nabla f(\bm{\theta}^t)}^2\notag\\
    &\leq {L\eta^2\tau^2(\lambda_k+1)}(2\kappa^2+\bar{\zeta}^2)\notag\\
    &\quad+ \frac{\eta L^2(2\gamma^2+1)}{4\gamma^2} \sum_{s=1}^{\tau} [16\eta^2\tau^2\gamma^2\left\| \nabla f(\bm{\theta}^t)\right\|^2 \notag\\
    &\quad+ 16\eta^2\tau^2\kappa^2+ {4\tau\eta^2}\bar{\zeta}^2] \notag\\
    & \quad+ \frac{\eta\tau\lambda_k\bar{\zeta}^2}{2\pi}-\frac{\eta\tau}{4} \mathbb{E}_t\left\|  \nabla f(\bm{\theta}^{t})\right\|^2  + \frac{Lk\sigma_0^2}{2r^3(\beta^t)^2}\label{in_conv_1}\\
    & = {L\eta^2\tau^2(\lambda_k+1)}(2\kappa^2+\bar{\zeta}^2) + \frac{Lk\sigma_0^2}{2r^3(\beta^t)^2}\notag\\
    &\quad+\frac{\eta^3\tau^2 L^2(2\gamma^2+1)(4\tau\kappa^2 + \bar{\zeta}^2)}{\gamma^2}  + \frac{\eta\tau\lambda_k\bar{\zeta}^2}{2\pi} \notag\\
    &\quad+\left(-\frac{\eta\tau}{4} + {4\eta^3\tau^3 L^2(2\gamma^2+1)} \right) \mathbb{E}_t\left\|\nabla f(\bm{\theta}^{t})\right\|^2\notag\\
    &  \leq \left({L\eta^2\tau^2}(2\kappa^2+\bar{\zeta}^2) + \frac{\eta\tau\bar{\zeta}^2}{2\pi} \right)\lambda_k +  \frac{Lk\sigma_0^2}{2r^3(\beta^t)^2}\notag\\
    &  \quad+  \eta^2\tau^2 L (12\eta\tau L + 2) \kappa^2 + \eta^2\tau^2 L (3\eta L + 1)\bar{\zeta}^2\notag\\
    &\quad-\frac{\eta\tau}{8}\mathbb{E}_t\left\|  \nabla f(\bm{\theta}^{t})\right\|^2\label{in_conv_2}\\
    & \leq -\frac{\eta\tau}{8}\mathbb{E}_t\left\|  \nabla f(\bm{\theta}^{t})\right\|^2 +  \eta^2\tau^2 L (3\kappa^2 + 2\bar{\zeta}^2) \notag\\
     &\quad+ \left({L\eta^2\tau^2}(2\kappa^2+\bar{\zeta}^2) + \frac{\eta\tau\bar{\zeta}^2}{2\pi} \right)\lambda_k  +  \frac{Lk\sigma_0^2}{2r^3(\beta^t)^2},\label{in_conv_3}
\end{align}
where~\eqref{in_conv_1} holds due to~\eqref{in_conv_eta1} and~\eqref{in_conv_eta2},~\eqref{in_conv_2} follows from~\eqref{in_conv_eta3} and $\gamma^2\geq1$, and~\eqref{in_conv_3} follows from~\eqref{in_conv_eta3}.

Rearranging the above inequality and summing it from $t=0$ to $T-1$, we get:
\begin{align}
\label{eqn:final_v2_topk}
    \sum_{t=0}^{T-1}\left\|  \nabla f(\bm{\theta}^{t})\right\|^2 \leq & \frac{8}{\eta\tau} \sum_{t=0}^{T-1}\mathbb{E}_t[f(\bm{\theta}^{t+1}) -f(\bm{\theta}^t)]  \notag\\
    &  +{8T}\left({\eta\tau L}(2\kappa^2+\bar{\zeta}^2) + \frac{\bar{\zeta}^2}{2\pi} \right)\lambda_k\notag\\
    & +8T\eta\tau L (3\kappa^2 + 2\bar{\zeta}^2)  \notag\\
    & +\sum_{t=0}^{T-1}\frac{4TLk\sigma_0^2}{\eta\tau r^3(\beta^t)^2},
\end{align}
where the expectation is taken over all rounds $t\in[0,T-1]$. Dividing both sides of \eqref{eqn:final_v2_topk} by $T$, one yields
\begin{align*}
    \frac{1}{T}\sum_{t=0}^{T-1}\left\|  \nabla f(\bm{\theta}^{t})\right\|^2  \leq &\frac{8(f(\bm{\theta}^{0}) -f_{\inf})}{T\eta\tau}   +  8\eta\tau L (3\kappa^2 + 2\bar{\zeta}^2)\\
    &+ {8}\left({\eta\tau L}(2\kappa^2+\bar{\zeta}^2)+\frac{\bar{\zeta}^2}{2\pi} \right)\lambda_k\notag\\
    &+ \frac{4Lk\sigma_0^2}{\eta\tau r^3}\sum_{t=0}^{T-1}\frac{1}{(\beta^t)^2}.
\end{align*}
By selecting a reasonable constant $\pi = 1/3$ (which satisfies $\pi < 1/2$), we arrive at the conclusion.
\end{IEEEproof}

We can see that the convergence bound~\eqref{in_theo_conv} contains three parts. The first part $8(f(\bm{\theta}^{0})-f^*)/(T\eta\tau)+8\eta\tau L(3\kappa^2+2\bar{\zeta}^2)$ is the \emph{optimization error} bound in FedAvg~\cite{bottou2018optimization}. The second part $8(\eta\tau L(2\kappa^2+\bar{\zeta}^2)+3\bar{\zeta}^2/2)\lambda_k/r$ is the \emph{compression error} resulting from applying $\randk_k$ to local model updates. When the $\randk_k$ is not applied, i.e., $k = d$ and $\lambda_k = 0$, the \emph{compression error} equals zero. The last part $4Lk/(\eta\tau r^3T)\sum_{t=0}^{T-1}\sigma_0^2/(\beta^t)^2$ is the \emph{privacy error}. When there is no privacy noise, i.e., $\sigma_0=0$, the \emph{privacy error} is equal to zero. Both privacy error and compression error raise the error floor at convergence. The last two terms explicitly show the trade-off between \emph{compression error} and \emph{privacy error} in terms of compression parameter $k$ in PFELS. As $k$ increases, $\lambda_k$ gets smaller, leading to a smaller compression error, but the privacy error increases. Therefore, in order to achieve the minimal convergence bound, we need to carefully choose an optimal parameter $k$ to balance these two errors.

\section{Convergence-Optimized Power Control under DP Guarantee }\label{sec_opt}
In this section, we develop a solution method to approximately solve $\mathbf{P1}$. First, we consider the power limit constraint~\eqref{eq_powe_cont} and power alignment constraint~\eqref{eq_pow_align}. From~\eqref{eq:spar_signal} and~\eqref{eq_pow_align}, the local transmission signal can be expressed as:
\begin{equation}
    \mathbf{x}_{i}^t = \frac{\beta^t}{|h_{i}^{t}|} \mathbf{A}^{t}\Delta_{i}^t, \quad \forall i\in\mathcal{S}^t, t.
\end{equation}
According to the power limit constraint $\mathbb{E} \norm{\mathbf{x}_i^t}_2^2 \leq P_i, \forall i, t$,  the power alignment coefficients $\{\beta^t\}_{t\in[T-1]}$ are constrained as follows:
\begin{equation}\label{powr_cstt}
    (\beta^t)^2\leq\min_{i\in \mathcal{S}^t}\frac{|h_{i}^{t}|^2P_i}{\mathbb{E}\norm{\mathbf{A}^t\Delta_i^t}_2^2},\quad \forall t.
\end{equation}
Here, to satisfy the power limit constraints, we can use the following lemma to approximate the power consumption $\mathbb{E}\norm{\mathbf{A}^t\Delta_i^t}_2^2$ under $\randk_k$ sparsification.
\begin{lemma}[Bounded Local Updates for $\randk_k$]\label{lemma_randk_powr} Under Assumption~\ref{ass_Bound_Grad}, given a local update $\Delta_{i}^{t}\in \mathbb{R}^d$ and $\randk_k$ sparsification with random projection matrix $\mathbf{A}^t$, we have
\begin{gather}\label{in_appx_ener}
\mathbb{E}\norm{\mathbf{A}^t\Delta_{i}^{t}}_{2}^{2} \leq \frac{k}{d}\eta^2\tau^2 C_{1}^2, \quad \forall i\in \mathcal{S}^t, t.
\end{gather}
\end{lemma}
\begin{IEEEproof}
The proof is provided in Appendix~\ref{append_proof_lemma_randk_powr} in the
supplementary.
\end{IEEEproof}

Then, we substitute the objective function and the DP constraint in \textbf{P1} with the convergence upper bound~\eqref{in_theo_conv} w.r.t. $\{\beta^t\}_{t\in[T-1]}$ in Theorem~\ref{theorem_conv} and the client-level DP result \eqref{eq_noise_low_bound} in Theorem~\ref{theo_dp_perRound}, respectively. Therefore, we can approximate Problem $\mathbf{P1}$ as follows:

\begin{subequations}\label{prob_opt_slot}
\begin{align}
    \mathbf{P2}\quad\min_{\{\beta^{t}\}_{t\in[T]} }&\quad\sum_{t=0}^{T-1}\frac{1}{(\beta^{t})^2}\label{prob_opt_slot_obj}\\
    \text{s.t.} & \quad C_2\beta^t\leq \epsilon, \quad \forall t \label{prob_opt_slot_dp_cs}\\
    & \quad 0<\beta^t\leq \min_{i\in \mathcal{S}^t}\frac{|h_i^t|\sqrt{dP_i}}{ C_1\eta\tau\sqrt{k}}, \quad \forall t.\label{prob_opt_slot_powr_cstt}
\end{align}
\end{subequations}
Note that $\mathbf{P2}$ can be readily solved as shown in the following result:
\begin{theorem}\label{theo_solu}
The optimal solution to Problem $\mathbf{P2}$ is given by:
\begin{equation}\label{eq_theo_opt_solu}
(\beta^t)^* = \min_{i\in\mathcal{S}^t}\{\frac{|h_i^t|\sqrt{dP_i}}{ C_1\eta\tau\sqrt{k}},\frac{\epsilon}{C_2}\},\quad \forall t. 
\end{equation}
\end{theorem}

\begin{IEEEproof} The objective~\eqref{prob_opt_slot_obj} is monotonically decreasing with respect to 
$\{\beta^t\}_{t\in[T-1]}$, and the constraints are decoupled across time. From the upper bounds of \eqref{prob_opt_slot_dp_cs} and \eqref{prob_opt_slot_powr_cstt}, we arrive at \eqref{eq_theo_opt_solu}.
\end{IEEEproof}

\section{Numerical Evaluation}
In this section, we conduct extensive experiments on common FL benchmark datasets to verify the performance of the proposed scheme. 

\subsection{Experimental Setup}

We consider a wireless FL system with 1000 devices and one central server. In each round, the server uniformly samples 32 devices to participate in the training process. To get a fair comparison with the baselines in the experimentation, we select the same MNIST-based dataset (FEMNIST) and CIFAR-10 dataset and corresponding model architectures in this paper as prior work in literature such as~\cite{hu2022federated,cheng2022differentially,geyer2017differentially,kairouz2021distributed,wei2021user}. Note that while FEMNIST and CIFAR-10 datasets are considered as ``solved'' in the computer vision community, achieving high accuracy with a strong DP guarantee remains difficult on these datasets~\cite{papernot2021tempered}.
CIFAR-10 is an image dataset that consists of 50,000 training images and 10,000 testing images. Each device has 50 training examples and 10 test examples by partitioning 50,000 samples over 1000 devices in an IID manner. Each image has a size of $32 \times 32$ pixels and an associated class label from 10 classes. The trained model on CIFAR-10 is a modified VGG-11 with 9,750,922 parameters in total. FEMNIST is the federated version of EMNIST dataset which has 3,550 users. Each image has a size of $28 \times 28$ pixels and an associated class label from 62 classes. We remove the users with less than 100 samples and randomly choose 1,000 users from the remaining set as all the participants. Each device has $90\%$ of images in the training set and $10\%$ in the test set. We train a modified ResNet-18~\cite{he2016deep} with 11,192,746 parameters in total on FEMNIST. 

For all experiments, we use mini-batch SGD with a momentum of 0.9 to train the local model with a batch size of 50. The learning rate of each algorithm is tuned from $\{0.01, 0.05, 0.1\}$ for CIAFR-10 and $\{0.001, 0.02, 0.05\}$ for FEMNIST using grid search. Following the implementation in \cite{reddi2020adaptive}, instead of doing $\tau$ local training steps per device, we perform $\tau$ epochs of training over each device's dataset. For CIFAR-10, the local epoch is set to $\tau=5$, and for FEMNIST, the local epoch is set to $\tau=5$. We tune the clipping threshold over the grid $C_1$ = \{1.0, 4.0, 8.0, 10.0\} and obtain the optimal clipping threshold, i.e., $C_1 = 1.0$ and $C_1=10.0$ for CIFAR-10 and FEMNIST, respectively. For all experiments, we set the privacy parameter $\delta=1/N$.

\begin{figure}[t]
\subfloat[CIFAR-10]{{\includegraphics[width=0.24\textwidth]{ {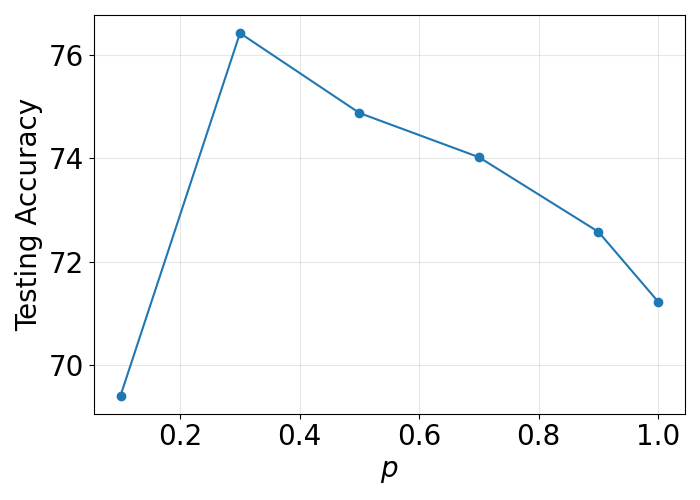}} }\label{fig:comp_acc_cifar}}
    \subfloat[FEMNIST]{{\includegraphics[width=0.24\textwidth]{ {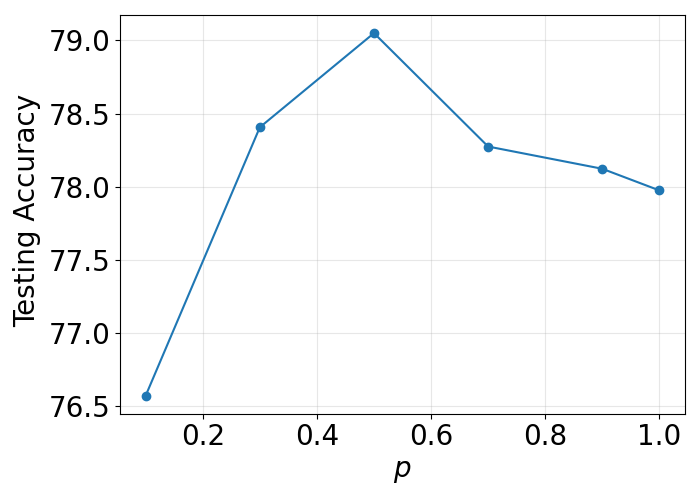}} }\label{fig:comp_acc_femnist}}
\caption{Test accuracy versus compression ratio $p$ for PFELS under $\epsilon=1.5$ for CIFAR-10 and $\epsilon=1.0$ for FEMNIST. 
}\label{fig:comp}
\end{figure}
We assume the channel gain $|h_i^t|$ follows an exponential distribution with a mean of $0.02$ and limit its value within the interval $[0.0001,0.1]$. The variance of the channel noise is set to $\sigma_0 = 1$. The maximum signal-to-noise ratio of each device $i$ is defined as $\textup{SNR}_i = P_i/d\sigma_0^2$. We set the maximum SNRs of all users to uniformly sampled from $2 \si{\dB}$ to $15 \si{\dB}$. We set the number of communication rounds $T=2000$ and $T=1000$ for CIFAR-10 and FEMNIST, respectively. Instead of using $k$, we define the \emph{compression ratio} $p := k/d$. We run each experiment with 5 random seeds and report the average. All algorithms are implemented using PyTorch on an Ubuntu server with 4 NVIDIA RTX 8000 GPUs. 

To show the effectiveness of PFELS, we simulate it under different constraints and compare it with several baselines. All baselines adopt the same uniform sampling strategy in each FL round.
\begin{itemize}
    \item WFL-P: \emph{Wireless FL without sparsification and DP constraint.} This baseline optimizes~\eqref{prob_opt_slot_obj} while transmitting the full model updates and ignoring the DP constraint~\eqref{prob_opt_slot_dp_cs}. According to Theorem~\ref{theo_solu}, when there are no requirements for compression and DP, the optimal power control decision is given by:
    \begin{equation}\label{eq_bas1}
       \beta^t = \min_{i\in\mathcal{S}^t}\{\frac{|h_i^t|\sqrt{P_i}}{C_1\eta\tau}\}, \quad\forall t. 
    \end{equation} 
    This baseline belongs to the standard AirComp-based FL algorithm and mimics the existing wireless FL method without privacy consideration \cite{sery2020analog}. 
    
    \item WFL-PDP: \emph{Wireless FL without sparsification and with DP constraint.} This baseline aims to optimize \eqref{prob_opt_slot_obj} while transmitting the full model update in each FL round and satisfying the DP constraint~\eqref{prob_opt_slot_dp_cs}. Similar to Theorem~\ref{theo_solu}, the optimal power control decision without sparsification is given by:
    \begin{equation}\label{eq_bas2}
        \beta^t= \min_{i\in\mathcal{S}^t}\{\frac{|h_i^t|\sqrt{P_i}}{C_1\eta\tau },\frac{\epsilon}{C_2}\}, \quad\forall t.
    \end{equation}
    This baseline mimics the state-of-the-art energy-efficient wireless FL method with privacy consideration \cite{koda2020differentially}. 

\end{itemize}

\subsection{Experimental Results}

\subsubsection{Impact of Compression Ratio in PFELS}
We first evaluate the impact of compression ratio on the performance of PFELS and show the results in Fig.~\ref{fig:comp} over both CIFAR-10 and FEMNIST datasets. From the figure, we can observe that the test accuracy of PFELS first increases and then decreases as the compression ratio $p$ increases from 0.1 to 1.0. This is consistent with the analysis in Theorem~\ref{theorem_conv}. Note that as $p$ increases, the \emph{compression error} decreases, but the \emph{privacy error} increases. Specifically, when the compression ratio is relatively small, i.e., $p=0.1$, the magnitude of \emph{compression error} is large and dominates the total convergence error, resulting in a higher training loss and lower testing accuracy. As $p$ increases, the \emph{compression error} reduces and leads to a lower loss and higher accuracy. However, when $p$ exceeds a threshold, i.e., $p=0.3$ for CIFAR-10 and $p=0.5$ for FEMNIST, the \emph{privacy error} becomes the dominant term in the total convergence error and keeps increasing as $p$ increases, resulting in a higher training loss and lower test accuracy. Therefore, one needs to carefully choose the optimal $p$ in practice to balance the privacy error and compression error. In the rest of the experiments, we always choose $p = 0.3$ for CIFAR-10 and $p = 0.5$ for FEMNIST in PFELS.

\subsubsection{Privacy-Accuracy Tradeoff in PFELS and Baselines}

\begin{figure}[t]

\subfloat[CIFAR-10]{{\includegraphics[width=0.24\textwidth]{ {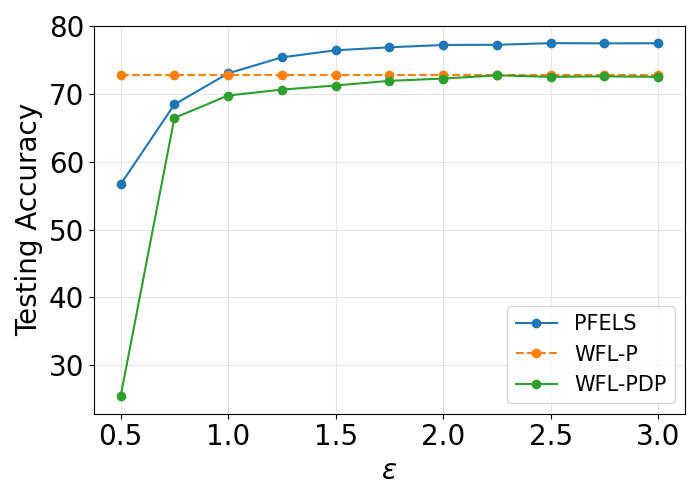}} }\label{fig:eps_acc_cifar}}
    \subfloat[FEMNIST]{{\includegraphics[width=0.24\textwidth]{ {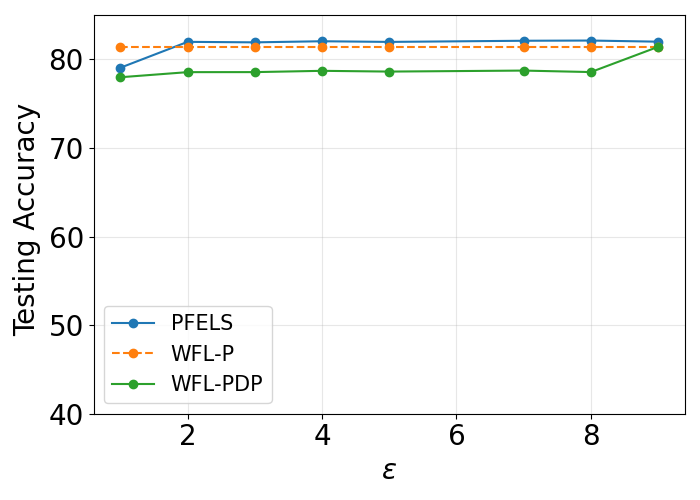}} }\label{fig:eps_loss_femnist}}
\caption{Test accuracy vs. DP privacy budget $\epsilon$ for PFELS and baselines.}\label{fig:EpsilonT}
\end{figure}

\begin{table}[!t]
\caption{Summary of results on CIFAR-10 dataset under $\epsilon=1.5$.\label{tab:cifar}}
\centering
\begin{tabular}{l|c|c|c}
\hline
Algorithm
 & Accuracy (\%) & Subcarriers ($d$)  & Energy cost (1e11) \\
\hline
 PFELS & 76.42\% & 600 & 2.55 \\
\hline
 WFL-P & 72.72\%& 2000 & 5.19 \\
\hline
 WFL-PDP & 72.23\%& 2000 & 3.70 \\
 \hline
\end{tabular}
\end{table}

\begin{table}[!t]
\caption{Summary of results on FEMNIST dataset under $\epsilon=2.0$.\label{tab:femnist}}
\centering
\begin{tabular}{l|c|c|c}
\hline
Algorithm
 & Accuracy (\%) & Subcarriers ($d$)  & Energy cost (1e11) \\
\hline
 PFELS & 81.97\% & 500 & 1.90 \\
\hline
 WFL-P & 81.43\%& 1000 &2.89 \\
\hline
 WFL-PDP & 78.55\%& 1000 &2.22 \\
 \hline
\end{tabular}
\end{table}

In this section, we compare the testing accuracies of PFELS and baselines by varying the privacy budget $\epsilon$ as shown in Fig.~\ref{fig:EpsilonT}. From the figure, we have the following observations. First, both the PFELS and WFL-PDP have a higher model accuracy as $\epsilon$ increases. This is due to the fact that a higher privacy budget $\epsilon$ indicates a lower noise requirement, and hence the useful signal has a relatively higher magnitude than the channel noise, leading to a more accurate model estimation in each round. Second, the accuracy of WFL-P is the upper bound of WFL-PDP because WFL-PDP needs to consider the additional DP constraint besides the same power constraint as in WFL-P. From \eqref{eq_bas1} and \eqref{eq_bas2}, we can see that the power control decisions of WFL-PDP and WFL-P are the same when $\epsilon$ is large. This is verified in Fig.~\ref{fig:EpsilonT}, which shows that the model accuracies of WFL-PDP and WFL-P are the same when $\epsilon\geq2.5$ for CIFAR-10 and $\epsilon\geq9$ for FEMNIST. 
Third, PFELS achieves is always better than WFL-PDP. This is due to PFELS sparsifies the local model updates before transmitting them to reduce the total convergence error while respecting the power and privacy constraints.
Fourth, when $\epsilon$ is relatively large (e.g., $\epsilon\geq 1.0$ for CIFAR-10 and $\epsilon\geq2.0$ for FEMNIST), leading to a non-binding DP constraint, PFELS can outperform WFL-P via the use of sparsification. In contrast, when $\epsilon$ is small, PFELS needs to adjust the power scaling coefficient to satisfy the strict DP constraint, while WFL-P lacks a DP constraint. Consequently, WFL-P can achieve higher model accuracy compared to PFELS in this case. 
Note PFELS sparsifies the local model updates before transmitting them to reduce the total convergence error while respecting the power and privacy constraints. This clearly shows the benefit of sparsification in improving the privacy-accuracy trade-off in wireless FL.

\subsubsection{Communication Efficiency Benefits of PFELS}
\begin{figure}[t]
    \subfloat[]{{\includegraphics[width=0.24\textwidth]{ {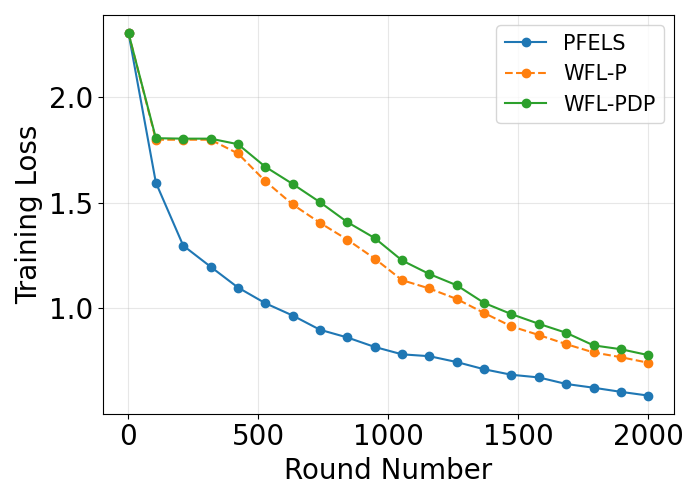}} }\label{fig:loss_round_cifar}}
    \subfloat[]{{\includegraphics[width=0.24\textwidth]{ {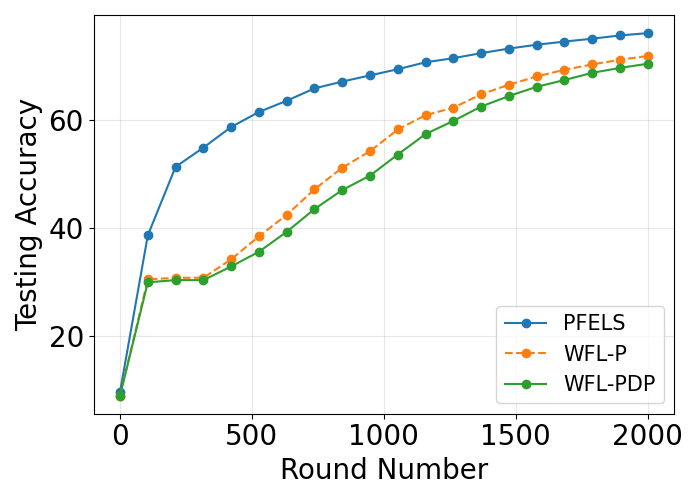}} }\label{fig:acc_round_cifar}}\\
    \subfloat[]{{\includegraphics[width=0.24\textwidth]{ {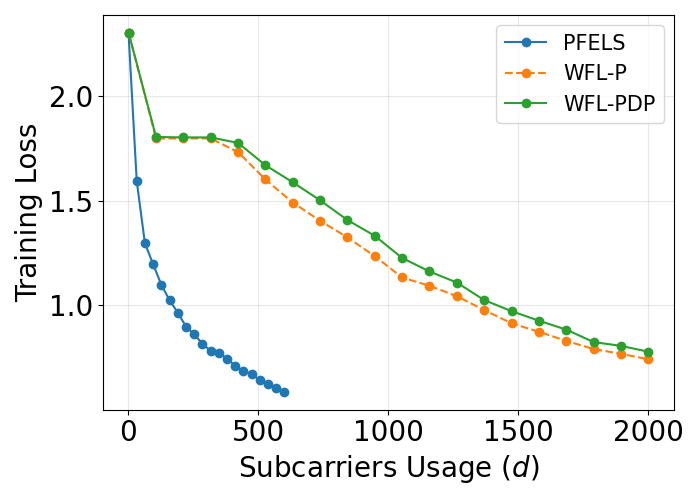}} }\label{fig:loss_chan_cifar}}
    \subfloat[]{{\includegraphics[width=0.24\textwidth]{ {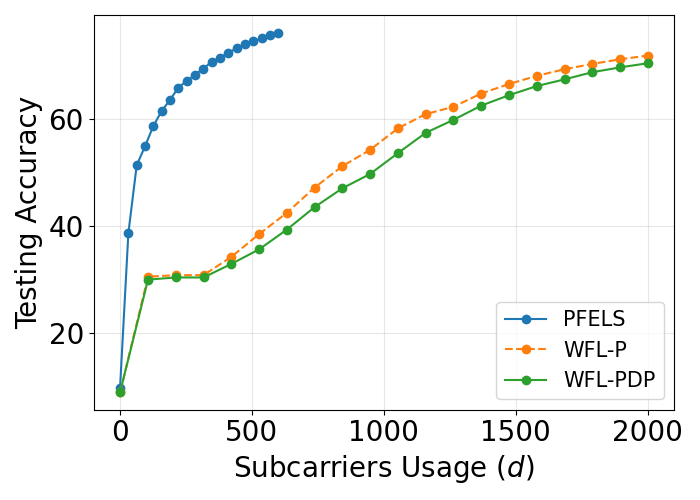}} }\label{fig:acc_chan_cifar}} 
    \caption{Training performance versus FL round and communication cost for CIFAR-10 under $\epsilon= 1.5$.}\label{fig:test_acc_round_cifar}
\end{figure}
\begin{figure}[t]
    \subfloat[]{{\includegraphics[width=0.24\textwidth]{ {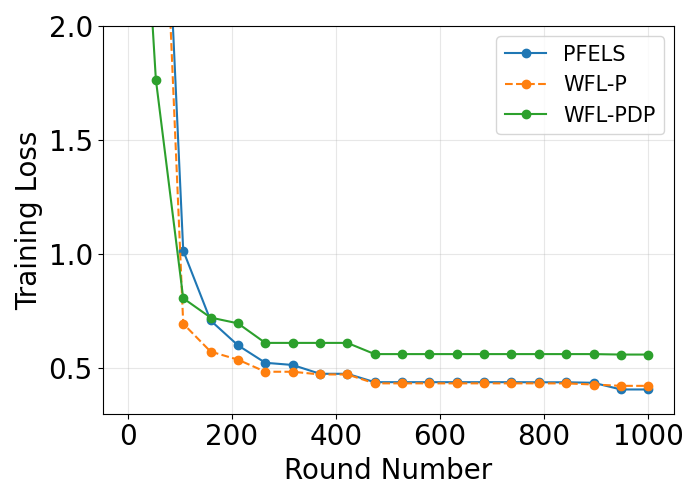}} }\label{fig:loss_round_femnist}}
    \subfloat[]{{\includegraphics[width=0.24\textwidth]{ {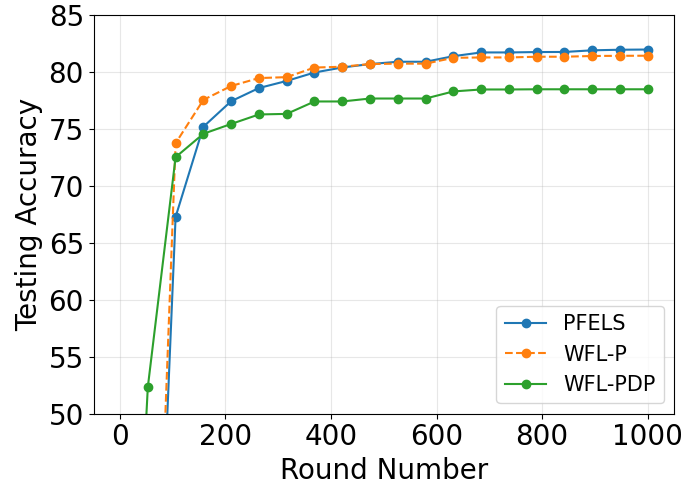}} }\label{fig:acc_round_femnist}}\\
    \subfloat[]{{\includegraphics[width=0.24\textwidth]{ {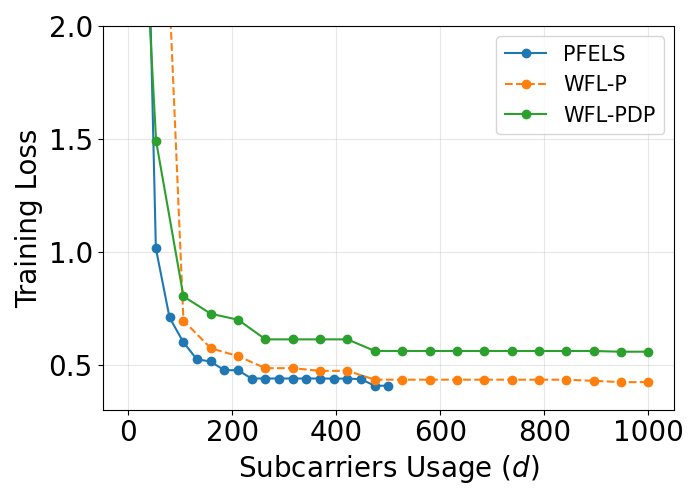}} }\label{fig:loss_chan_femnist}} 
    \subfloat[]{{\includegraphics[width=0.24\textwidth]{ {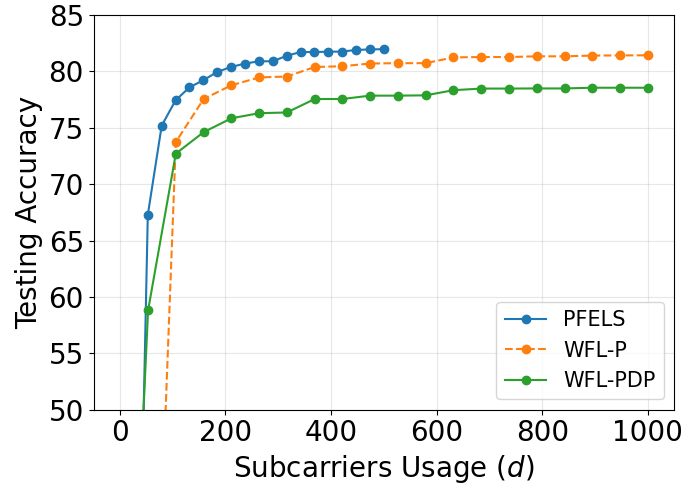}} }\label{fig:acc_chan_femnist}}
    \caption{Training performance versus FL round and communication cost for FEMNIST under $\epsilon= 2.0$.}\label{fig:test_acc_round_femnist}
\end{figure}

Next, we compare the convergence rate and communication cost in terms of subcarrier usage for PFELS and baselines under a fixed privacy budget. Table~\ref{tab:cifar} and Table~\ref{tab:femnist} summarize the performance of PFELS and baselines after $T$ rounds on CIFAR-10 and FEMNIST, respectively. Fig.~\ref{fig:test_acc_round_cifar} and Fig.~\ref{fig:test_acc_round_femnist} show the performance during the training on CIFAR-10 and FEMNIST, respectively. First, for CIFAR-10, as shown in Fig.~\ref{fig:loss_round_cifar} on training loss and Fig.~\ref{fig:acc_round_cifar} on test accuracy, PFELS has the fastest convergence speed in terms of communication round. This is due to the fact that PFELS uses compression to balance the trade-off between the privacy error and compression error and achieves a better convergence speed. Second, WFL-P has a faster convergence speed in terms of communication round than WFL-PDP because WFL-PDP needs to consider the additional DP constraint besides the same power constraint as in WFL-P. Third, Fig.~\ref{fig:loss_chan_cifar} and Fig.~\ref{fig:acc_chan_cifar} show the training loss and test accuracy w.r.t. the communication cost for PFELS and baselines. The results in Table~\ref{tab:cifar} show that PFELS achieves higher communication efficiency than the baselines by utilizing sparsification with only 600 subcarriers. In contrast, both WFL-P and WFL-PDP incur the same communication cost, using 2000 subcarriers, as they transmit complete model updates. Similar results can be observed for FEMNIST dataset from Fig.~\ref{fig:test_acc_round_femnist} and Table~\ref{tab:femnist}.

\subsubsection{Energy Efficiency Benefits of PFELS}
\begin{figure}[t]
    \subfloat[CIFAR-10]{{\includegraphics[width=0.24\textwidth]{ {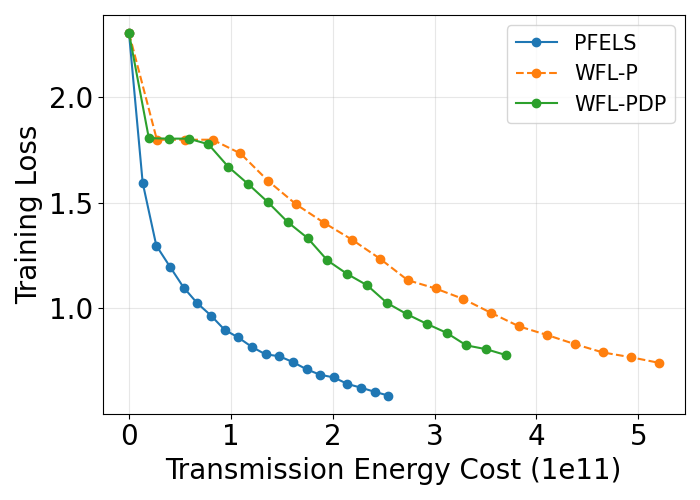}} }\label{fig:loss_power_cifar}}
    \subfloat[CIFAR-10]{{\includegraphics[width=0.24\textwidth]{ {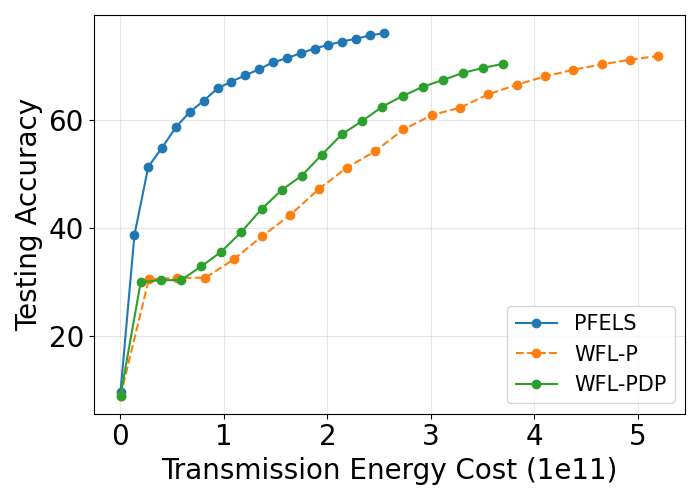}} }\label{fig:acc_power_cifar}}\\
    \subfloat[FEMNIST]{{\includegraphics[width=0.24\textwidth]{ {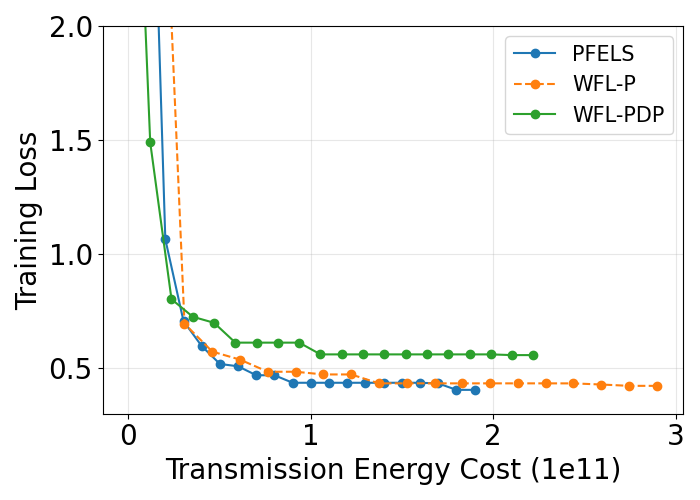}} }\label{fig:loss_power_femnist}}
    \subfloat[FEMNIST]{{\includegraphics[width=0.24\textwidth]{ {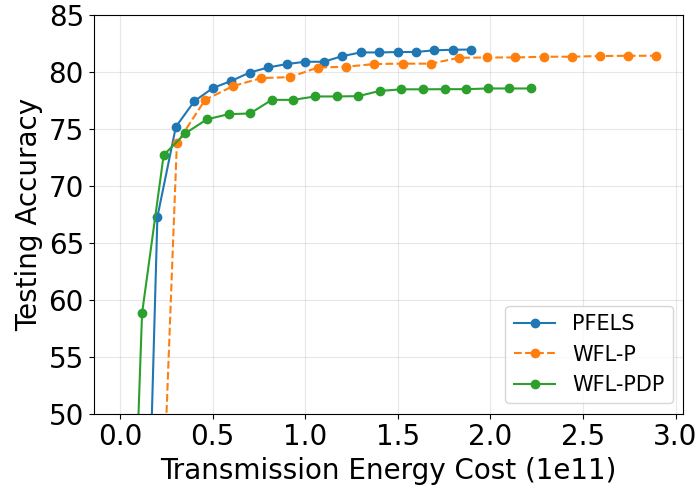}} }\label{fig:acc_power_femnist}}
    \caption{Transmission energy cost comparisons for CIFAR-10 under $\epsilon= 1.5$ and FEMNIST datasets under $\epsilon= 2.0$.}\label{fig:energy}
\end{figure}

Finally, we evaluate the benefit of PFELS in improving communication energy efficiency compared with the baselines under a fixed privacy budget. Here, we compute the total of accumulated transmission energy of all participated devices as the transmission energy cost. Table~\ref{tab:cifar},~\ref{tab:femnist}, and Fig.~\ref{fig:energy} show the results on CIFAR-10 and FEMNIST. For CIFAR-10, as shown in Fig.~\ref{fig:loss_power_cifar},~\ref{fig:acc_power_cifar}, and Table~\ref{tab:cifar}, we observe that PFELS achieves the final test accuracy (76.42\%) while utilizing $2.55\times10^{11}\si{J}$ in energy consumption, whereas WFL-P consumes $5.19\times10^{11}\si{J}$ to achieve the final test accuracy (72.72\%) and WFL-P consumes $3.70\times10^{11}\si{J}$ to achieve the final test accuracy (72.23\%) in energy consumption.
For FEMNIST dataset, as depicted in Fig.~\ref{fig:loss_power_femnist},~\ref{fig:acc_power_femnist}, and Table~\ref{tab:femnist}, PFELS achieves the final test accuracy (81.97\%) while utilizing $1.90\times10^{11}\si{J}$ in energy consumption, whereas WFL-P consumes $2.89\times10^{11}\si{J}$ to achieve the final test accuracy (81.43\%) and WFL-P consumes $2.22\times10^{11}\si{J}$ to achieve the final test accuracy (78.55\%) in energy consumption.
In summary, PFELS outperforms all baselines by saving transmission energy cost.

\section{Conclusion}
In this paper, we have proposed PFELS, a novel wireless FL scheme aimed at achieving client-level DP while maintaining high model accuracy and improving communication and energy efficiency by harnessing the intrinsic channel noise, signal-superposition nature of the wireless channel, and update compression in AirComp. Furthermore, we have analyzed the convergence and privacy properties of PFELS under the general non-convex and non-iid setting. Experimental results have demonstrated that compared with prior wireless FL frameworks, PFELS can greatly improve the model accuracy, communication and energy efficiency simultaneously under the same DP guarantee. In the future, we will investigate other compression methods and extend our algorithm to the setting with imperfect channel status. 

\ifCLASSOPTIONcompsoc
  \section*{Acknowledgments}
\else
  \section*{Acknowledgment}
\fi
The work of Z. Zhang, Y. Guo, and Y. Gong was partially supported by NSF under grants CNS-2047761, CNS-2106761, CMMI-2222670, and CNS-2318683. 

\bibliographystyle{IEEEtran}

\bibliography{main_arxiv.bib}

\ifCLASSOPTIONcaptionsoff
  \newpage
\fi

\newpage
\onecolumn
\title{Scalable and Low-Latency Federated Learning with Cooperative Mobile Edge Networking (Supplementary)}

\appendices
\section{Proof of Lemma 1}\label{append_proof_lemma_estimatationp}
\begin{IEEEproof}
For the $\randk_k$, taking the expectation on the active set $\omega$, we have:
\begin{subequations}
\begin{align}
    \mathbb{E}_{\omega}[\hat{\Delta}^t] & =
    \mathbb{E}_{\omega}[\frac{1}{r}\sum_{i\in\mathcal{S}^t} (\mathbf{A}^{t})^{\intercal}\mathbf{A}^{t}\Delta_{i }^t + \frac{(\mathbf{A}^{t})^{\intercal}\mathbf{z}^{t}}{r\beta^t}] \\
    & = \frac{1}{r}\sum_{i\in\mathcal{S}^t}\mathbb{E}_{\omega}[ (\mathbf{A}^{t})^{\intercal}\mathbf{A}^{t}\Delta_{i }^t]\label{eq_delta_hat}\\
    & = \frac{1}{r}\sum_{i\in\mathcal{S}^t}[\frac{k}{d}[\Delta_{i}^t]_1,\dots,\frac{k}{d}[\Delta_{i}^t]_d]\\
    & = \frac{k}{d}\sum_{i\in\mathcal{S}^t}\frac{\Delta_{i}^t}{r},
\end{align}
\end{subequations}
where~\eqref{eq_delta_hat} holds due to the zero mean of Gaussian noise.
\end{IEEEproof}

\section{Proof of Lemma 3}\label{append_proof_iteate_decomp}
\begin{IEEEproof}
According to the Assumption \ref{ass:smoothness}, we have:
\begin{align}
    \mathbb{E}_t & [f(\bm{\theta}^{t+1}-f(\bm{\theta}^t)] \leq   \mathbb{E}_t\langle \nabla f(\bm{\theta}^t),\bm{\theta}^{t+1}-\bm{\theta}^t\rangle+\frac{L}{2}\mathbb{E}_t\norm{\bm{\theta}^{t+1}-\bm{\theta}^t}^2\notag\\
    & = \mathbb{E}_t\langle \nabla f(\bm{\theta}^t),\mathbb{E}_{\mathcal{S}^t}\left[\frac{1}{r}\sum_{i\in\mathcal{S}^t}(\mathbf{A}^t)^\intercal\mathbf{A}^t\Delta_{i}^{t}+\frac{(\mathbf{A}^t)^\intercal\mathbf{z}^t}{r\beta^t}\right]\rangle+\frac{L}{2}\mathbb{E}_t\norm{\frac{1}{r}\sum_{i\in\mathcal{S}^t}(\mathbf{A}^t)^\intercal\mathbf{A}^t\Delta_{i}^{t}+\frac{(\mathbf{A}^t)^\intercal\mathbf{z}^t}{r\beta^t}}^2\notag\\
    & \labelrel={eq_model_update1} \mathbb{E}_t\langle \nabla f(\bm{\theta}^t),\mathbb{E}_{\mathcal{S}^t}\left[\frac{1}{r}\sum_{i\in\mathcal{S}^t}(\mathbf{A}^t)^\intercal\mathbf{A}^t\Delta_{i}^{t}\right]\rangle+\frac{L}{2}\mathbb{E}_t\norm{\frac{1}{r}\sum_{i\in\mathcal{S}^t}(\mathbf{A}^t)^\intercal\mathbf{A}^t\Delta_{i}^{t}+\frac{(\mathbf{A}^t)^\intercal\mathbf{z}^t}{r\beta^t}}^2\notag\\
    & \labelrel={eq_model_update2} \mathbb{E}_t\langle \nabla f(\bm{\theta}^t), \frac{1}{r}r\frac{1}{N}\sum_{i=1}^N(\mathbf{A}^t)^\intercal\mathbf{A}^t\Delta_{i}^{t}\rangle+\frac{L}{2}\mathbb{E}_t\norm{\frac{1}{r}\sum_{i\in\mathcal{S}^t}(\mathbf{A}^t)^\intercal\mathbf{A}^t\Delta_{i}^{t}+\frac{(\mathbf{A}^t)^\intercal\mathbf{z}^t}{r\beta^t}}^2\notag\\
    & = \underbrace{\mathbb{E}_t\langle \nabla f(\bm{\theta}^t), \frac{1}{N}\sum_{i=1}^N(\mathbf{A}^t)^\intercal\mathbf{A}^t\Delta_{i}^{t}\rangle}_{T_1}+\underbrace{\frac{L}{2}\mathbb{E}_t\norm{\frac{1}{r}\sum_{i\in\mathcal{S}^t}(\mathbf{A}^t)^\intercal\mathbf{A}^t\Delta_{i}^{t}+\frac{(\mathbf{A}^t)^\intercal\mathbf{z}^t}{r\beta^t}}^2}_{T_2}\label{eq_t1_t2}
\end{align}
where~\eqref{eq_model_update1} holds due to the zero-mean Gaussian noise,~\eqref{eq_model_update2} follows from the unbiased sampling.
For $T_1$, we have:
\begin{align}
    T_1 & = \mathbb{E}_t\langle \nabla f(\bm{\theta}^t), \frac{1}{N}\sum_{i=1}^N(\mathbf{A}^t)^\intercal\mathbf{A}^t\Delta_{i}^{t}-\frac{1}{N}\sum_{i=1}^N{\Delta_{i}^{t}}+\frac{1}{N}\sum_{i=1}^N{\Delta_{i}^{t}}\rangle\notag\\
    & = \underbrace{\mathbb{E}_t\langle \nabla f(\bm{\theta}^t), \frac{1}{N}\sum_{i=1}^N{\Delta_{i}^{t}}\rangle}_{A_1} + \underbrace{\mathbb{E}_t\langle \nabla f(\bm{\theta}^t), \frac{1}{N}\sum_{i=1}^N(\mathbf{A}^t)^\intercal\mathbf{A}^t\Delta_{i}^{t}-\frac{1}{N}\sum_{i=1}^N{\Delta_{i}^{t}}\rangle}_{A_2}\label{t1_a1_a2}
\end{align}
For $A_1$, we have:
\begin{align}
    A_1 & = \mathbb{E}_t\langle \nabla f(\bm{\theta}^t), \frac{1}{N}\sum_{i=1}^N\Delta_{i}^{t}\rangle\notag\\
    & \labelrel={eq_t1_upd} -\eta\tau\mathbb{E}_t\langle \nabla f(\bm{\theta}^t),\frac{1}{N\tau}\sum_{i=1 }^{N} \sum_{s=1}^{\tau} \nabla f_i(\bm{\theta}_i^{t,s-1})\rangle\notag\\
    & \labelrel={eq_t1_square_diff} -\frac{\eta\tau }{2}\norm{\nabla f(\bm{\theta}^t)}^2 
    -\frac{\eta\tau }{2}\norm{\frac{1}{N\tau}\sum_{i=1 }^{N} \sum_{s=1}^{\tau} \nabla f_i(\bm{\theta}_i^{t,s-1})}^2 
    + \frac{\eta\tau }{2}\norm{\nabla f(\bm{\theta}^t)-\frac{1}{N\tau}\sum_{i=1 }^{N} \sum_{s=1}^{\tau} \nabla f_i(\bm{\theta}_i^{t,s-1})}^2\notag\\
    & = -\frac{\eta\tau}{2}\norm{\nabla f(\bm{\theta}^t)}^2    
    + \frac{\eta\tau }{2}\norm{\frac{1}{N\tau}\sum_{i=1}^{N} \sum_{s=1}^{\tau}(\nabla f_i(\bm{\theta}^t)- \nabla f_i(\bm{\theta}_i^{t,s-1}))}^2-\frac{\eta\tau }{2}\norm{\frac{1}{N\tau}\sum_{i=1 }^{N} \sum_{s=1}^{\tau} \nabla f_i(\bm{\theta}_i^{t,s-1})}^2\notag\\
    & \labelrel\leq{in_t1_jen} -\frac{\eta\tau }{2}\norm{\nabla f(\bm{\theta}^t)}^2
    + \frac{\eta }{2N}\sum_{i=1 }^{N} \sum_{s=1}^{\tau}\norm{\nabla f_i(\bm{\theta}^t)- \nabla f_i(\bm{\theta}_i^{t,s-1})}^2-\frac{\eta\tau }{2}\norm{\frac{1}{N\tau}\sum_{i=1 }^{N} \sum_{s=1}^{\tau} \nabla f_i(\bm{\theta}_i^{t,s-1})}^2\notag\\
    & \labelrel\leq{in_t1_smooth} -\frac{\eta\tau}{2}\norm{\nabla f(\bm{\theta}^t)}^2+\frac{\eta L^2}{2N}\sum_{i=1 }^{N} \sum_{s=1}^{\tau}\norm{\bm{\theta}^t- \bm{\theta}_i^{t,s-1}}^2-\frac{\eta\tau }{2}\norm{\frac{1}{N\tau}\sum_{i=1 }^{N} \sum_{s=1}^{\tau} \nabla f_i(\bm{\theta}_i^{t,s-1})}^2,\label{a1}
\end{align}
where~\eqref{eq_t1_upd} holds due to $ \bm{\theta}_{i}^{t,\tau} = \bm{\theta}^{t}-\eta\sum_{s=1}^{\tau}\mathbf{g}_{i}^{t,s-1}$ and the unbiased stochastic gradient Assumption \ref{ass:gradient}, \eqref{eq_t1_square_diff} follows from $2\langle \bm{a},\bm{b}\rangle = \norm{\bm{a}}^2+\norm{\bm{b}}^2-\norm{\bm{a}-\bm{b}}^2$,~\eqref{in_t1_jen} holds due to Jensen's inequality, and~\eqref{in_t1_smooth} comes from the $L$-smoothness Assumption \ref{ass:smoothness}.

For $A_2$, let $\lambda_k=1-\frac{k}{d}$, we have:
\begin{align}
    A_2 & \labelrel={eq_a2_upd} \eta\tau\mathbb{E}_t\langle \nabla f(\bm{\theta}^t), \frac{1}{N\tau}\sum_{i=1}^N\sum_{s=1}^{\tau}\mathbf{g}_{i}^{t,s-1} - \frac{1}{N\tau}\sum_{i=1}^N\sum_{s=1}^{\tau}(\mathbf{A}^t)^\intercal\mathbf{A}^t\mathbf{g}_{i}^{t,s-1}\rangle\notag\\
    & \labelrel\leq{in_a2_lemma_gamma} \frac{\eta\tau}{2}\left( \pi\norm{\nabla f(\bm{\theta}^t)}^2 + \pi^{-1}\mathbb{E}_t\norm{\frac{1}{N\tau}\sum_{i=1}^N\sum_{s=1}^{\tau}\mathbf{g}_{i}^{t,s-1} - \frac{1}{N\tau}\sum_{i=1}^N\sum_{s=1}^{\tau}(\mathbf{A}^t)^\intercal\mathbf{A}^t\mathbf{g}_{i}^{t,s-1}}^2\right)\notag\\
    & \labelrel\leq{in_a2_bound_spar} \frac{\eta\tau\pi}{2}\norm{\nabla f(\bm{\theta}^t)}^2 + \frac{\eta\tau\lambda_k}{2\pi}\mathbb{E}_t\norm{\frac{1}{N\tau}\sum_{i=1}^N\sum_{s=1}^{\tau}\mathbf{g}_{i}^{t,s-1}}^2\notag\\
    & = \frac{\eta\tau\pi}{2}\norm{\nabla f(\bm{\theta}^t)}^2 + \frac{\eta\tau\lambda_k}{2\pi}\mathbb{E}_t\norm{\frac{1}{N\tau}\sum_{i=1}^N\sum_{s=1}^{\tau}\mathbf{g}_{i}^{t,s-1} - \frac{1}{N\tau}\sum_{i=1}^N\sum_{s=1}^{\tau}\nabla f_i(\bm{\theta}_i^{t,s-1}) + \frac{1}{N\tau}\sum_{i=1}^N\sum_{s=1}^{\tau}\nabla f_i(\bm{\theta}_i^{t,s-1})}^2\notag\\
    & \labelrel={eq_a2_cross} \frac{\eta\tau\pi}{2}\norm{\nabla f(\bm{\theta}^t)}^2 + \frac{\eta\tau\lambda_k}{2\pi}\mathbb{E}_t\left[\norm{\frac{1}{N\tau}\sum_{i=1}^N\sum_{s=1}^{\tau}\mathbf{g}_{i}^{t,s-1} - \frac{1}{N\tau}\sum_{i=1}^N\sum_{s=1}^{\tau}\nabla f_i(\bm{\theta}_i^{t,s-1})}^2 + \norm{\frac{1}{N\tau}\sum_{i=1}^N\sum_{s=1}^{\tau}\nabla f_i(\bm{\theta}_i^{t,s-1})}^2\right]\notag\\
    & \labelrel\leq{in_a2_jens} \frac{\eta\tau\pi}{2}\norm{\nabla f(\bm{\theta}^t)}^2 + \frac{\eta\tau\lambda_k}{2N\pi}\sum_{i=1}^N\mathbb{E}_t\norm{\frac{1}{\tau}\sum_{s=1}^{\tau}\mathbf{g}_{i}^{t,s-1} - \frac{1}{\tau}\sum_{s=1}^{\tau}\nabla f_i(\bm{\theta}_i^{t,s-1})}^2 + \frac{\eta\tau\lambda_k}{2\pi}\mathbb{E}_t\norm{\frac{1}{N\tau}\sum_{i=1}^N\sum_{s=1}^{\tau}\nabla f_i(\bm{\theta}_i^{t,s-1})}^2\notag\\
    & \labelrel\leq{in_a2_bound_dis} \frac{\eta\tau\pi}{2}\norm{\nabla f(\bm{\theta}^t)}^2 + \frac{\eta\tau\lambda_k\bar{\zeta}^2}{2\pi} + \frac{\eta\tau\lambda_k}{2\pi}\mathbb{E}_t\norm{\frac{1}{N\tau}\sum_{i=1}^N\sum_{s=1}^{\tau}\nabla f_i(\bm{\theta}_i^{t,s-1})}^2\label{a2}
\end{align}
where~\eqref{eq_a2_upd} holds due to $ \bm{\theta}_{i}^{t,\tau} = \bm{\theta}^{t}-\eta\sum_{s=1}^{\tau}\mathbf{g}_{i}^{t,s-1}$,~\eqref{in_a2_lemma_gamma} follows from Lemma~\ref{lemma_in_inn_prod},~\eqref{in_a2_bound_spar} follows from Lemma~\ref{lemma_bound_spar},~\eqref{eq_a2_cross} uses unbiased Assumption~\ref{ass:gradient},~\eqref{in_a2_jens} uses Lemma~\ref{lemma_jensen_in}, and~\eqref{in_a2_bound_dis} uses~\eqref{in_sgd_gd_jens} and $\bar{\zeta}^2=(1/N)\sum_{i=1}^{N}\zeta_{i}^{2}$.

Combining~\eqref{t1_a1_a2},~\eqref{a1} and~\eqref{a2}, if $\lambda_k\leq\pi$, we get:
\begin{align}
    T_1 & \leq -\frac{\eta\tau (1- \pi)}{2}\left\|  \nabla f(\bm{\theta}^{t})\right\|^2 + \frac{\eta L^2}{2N} \sum_{i=1}^{N}\sum_{s=1}^{\tau} \mathbb{E}_t\left\| \bm{\theta}^{t}- \bm{\theta}_i^{t,s-1} \right\|^2  + \frac{\eta\tau\lambda_k\bar{\zeta}^2}{2\pi}- \frac{\eta\tau (1-\lambda_k/\pi)}{2}\mathbb{E}_t\norm{\frac{1}{N\tau}\sum_{i=1}^N\sum_{s=1}^{\tau}\nabla f_i(\bm{\theta}_i^{t,s-1})}^2\notag\\
    & \leq  -\frac{\eta\tau (1- \pi)}{2}\left\|  \nabla f(\bm{\theta}^{t})\right\|^2 + \frac{\eta L^2}{2N} \sum_{i=1}^{N}\sum_{s=1}^{\tau} \mathbb{E}_t\left\| \bm{\theta}^{t}- \bm{\theta}_i^{t,s-1} \right\|^2  + \frac{\eta\tau\lambda_k\bar{\zeta}^2}{2\pi}.\label{in_t1}
\end{align}
For $T_2$, using lemma because the expectation of Gaussian noise is zero, we have:
\begin{align*}
    T_2  & = \frac{L}{2}\mathbb{E}_t\norm{\frac{1}{r}\sum_{i\in\mathcal{S}^t}(\mathbf{A}^t)^\intercal\mathbf{A}^t\Delta_{i}^{t} + \frac{(\mathbf{A}^t)^\intercal\mathbf{z}^t}{r\beta^t}}^2\notag\\
    & \labelrel={eq_t2_indepent} \frac{L}{2}\mathbb{E}_t\norm{\frac{1}{r}\sum_{i\in \mathcal{S}_t} (\mathbf{A}^t)^\intercal\mathbf{A}^t\Delta_{i}^{t}}^2 
    + \frac{L}{2}\mathbb{E}_t\norm{\frac{1}{r}\sum_{i\in\mathcal{S}^t} \frac{(\mathbf{A}^t)^\intercal\mathbf{z}^t}{r\beta^t}}^2\notag\\
    & = \frac{L}{2}\mathbb{E}_t\norm{\frac{1}{r}\sum_{i\in\mathcal{S}^t}(\mathbf{A}^t)^\intercal\mathbf{A}^t\Delta_{i}^{t} -\frac{1}{r}\sum_{i\in\mathcal{S}^t}\Delta_{i}^{t} + \frac{1}{r}\sum_{i\in\mathcal{S}^t}\Delta_{i}^{t}}^2 
    + \frac{L}{2r^2}r\frac{\sigma_0^2k}{(r\beta^t)^2}\\
    & \labelrel\leq{in_t2_square} \frac{L}{r^2}\mathbb{E}_t\norm{\sum_{i\in\mathcal{S}^t}(\mathbf{A}^t)^\intercal\mathbf{A}^t\Delta_{i}^{t}-\Delta_{i}^{t}}^2 
    + \frac{L}{r^2}\mathbb{E}_t\norm{\sum_{i\in\mathcal{S}^t}\Delta_{i}^{t}}^2 
    + \frac{Lk\sigma_0^2}{2r^3(\beta^t)^2}\\
    & \labelrel\leq{in_t2_sum_square} \frac{L}{r^2}r\mathbb{E}_t\left[\sum_{i\in \mathcal{S}^t}\norm{(\mathbf{A}^t)^\intercal\mathbf{A}^t\Delta_{i}^{t}-\Delta_{i}^{t}}^2\right] 
    + \frac{L}{r^2}r\mathbb{E}_t\left[\sum_{i\in\mathcal{S}^t}\norm{\Delta_{i}^{t}}^2\right]
    + \frac{Lk\sigma_0^2}{2r^3(\beta^t)^2}\\
    & \labelrel={eq_t2_var_randk}\frac{L}{r}\lambda_k\mathbb{E}_t\left[\sum_{i\in\mathcal{S}^t}\norm{\Delta_{i}^{t}}^2\right] 
    + \frac{L}{r}\mathbb{E}_t\left[\sum_{i\in\mathcal{S}^t}\norm{\Delta_{i}^{t}}^2\right]
    + \frac{Lk\sigma_0^2}{2r^3(\beta^t)^2}\\
    & = L(\lambda_k+1)\frac{1}{N}\sum_{i=1}^{N}\mathbb{E}_t\norm{\Delta_{i}^{t}}^2
    +\frac{Lk\sigma_0^2}{2r^3(\beta^t)^2}\\
    & \labelrel={eq_t2_upd}  \frac{L\eta^2\tau^2}{N}(\lambda_k+1)\sum_{i=1}^{N}\mathbb{E}_t\norm{\frac{1}{\tau}\sum_{s=1}^{\tau}\mathbf{g}_{i}^{t,s-1}}^2
    +\frac{Lk\sigma_0^2}{2r^3(\beta^t)^2},
\end{align*}
where \eqref{eq_t2_indepent} holds due to the fact that $\mathbf{z}^t$ is zero-mean,~\eqref{in_t2_square} and ~\eqref{in_t2_sum_square} follow from Lemma~\ref{lemma_cau_sch_in},~\eqref{eq_t2_var_randk} follows from Lemma~\ref{lemma_bound_spar}, and~\eqref{eq_t2_upd} holds due to $ \bm{\theta}_{i}^{t,\tau} = \bm{\theta}^{t}-\eta\sum_{s=1}^{\tau}\mathbf{g}_{i}^{t,s-1}$.

According to Lemma \ref{lemma_bound_loc_ups}, we have:
\begin{align}
     T_2  \leq & 2L\eta^2\tau^2\gamma^2(\lambda_k+1)\norm{\nabla f(\bm{\theta}^t)}^2
     +\frac{2L^3\eta^2\tau}{N}(\lambda_k+1)\sum_{i=1}^{N}\sum_{s=1}^{\tau}\mathbb{E}_t\norm{\bm{\theta}^t-\bm{\theta}_{i}^{t,s-1}}^2\notag\\
     &+L\eta^2\tau^2(\lambda_k+1)(\bar{\zeta}^2+2\kappa^2)
      + \frac{Lk\sigma_0^2}{2r^3(\beta^t)^2} \label{t2_rand_final}
\end{align}
Combining \eqref{eq_t1_t2}, \eqref{in_t1} and \eqref{t2_rand_final}, we get:
\begin{align*}
     \mathbb{E}_t  [f(\bm{\theta}^{t+1}-f(\bm{\theta}^t)] \leq & \left( -\frac{\eta\tau (1-\pi)}{2}+2L\eta^2\tau^2\gamma^2(\lambda_k+1) \right)\norm{\nabla f(\bm{\theta}^t)}^2 \\
     & + \left( \frac{\eta L^2}{2N} + \frac{2L^3\eta^2\tau^2(\lambda_k+1)}{n\tau}\right)\sum_{i=1}^{N}\sum_{s=1}^{\tau}\mathbb{E}_t\norm{\bm{\theta}^t-\bm{\theta}_{i}^{t,s-1}}^2 + \frac{\eta\tau\lambda_k\bar{\zeta}^2}{2\pi}\\
     & +L\eta^2\tau^2(\lambda_k+1)(\bar{\zeta}^2+2\kappa^2) + \frac{Lk\sigma_0^2}{2r^3(\beta^t)^2}.
\end{align*}
\end{IEEEproof}

\section{Proof of lemma 4}\label{append_proof_bound_locl_divg}
\begin{IEEEproof}
According to the local update rule, we have
\begin{align*}
    \mathbb{E}_t&\norm{\bm{\theta}^t-\bm{\theta}_{i}^{t,s-1}}^2 =  \mathbb{E}_t\norm{\bm{\theta}_{i}^{t,s-2}-\bm{\theta}^t-\eta\mathbf{g}_{i}^{t,s-2}}^2 \notag\\
    & = \mathbb{E}_t\norm{\bm{\theta}_{i}^{t,s-2}-\bm{\theta}^t-\eta\mathbf{g}_{i}^{t,s-2}+\eta\nabla f_i(\bm{\theta}_{i}^{t,s-2})-\eta\nabla f_i(\bm{\theta}_{i}^{t,s-2})+\eta\nabla f_i(\bm{\theta}^{t})-\eta\nabla f_i(\bm{\theta}^{t})}^2\notag\\
    & \labelrel={eq_ld_cross} \mathbb{E}_t\norm{\bm{\theta}_{i}^{t,s-2}-\bm{\theta}^t-\eta\nabla f_i(\bm{\theta}_{i}^{t,s-2})+\eta\nabla f_i(\bm{\theta}^{t})-\eta\nabla f_i(\bm{\theta}^{t})}^2 
    +\eta^2\mathbb{E}_t\norm{\mathbf{g}_{i}^{t,s-2}-\nabla f_i(\bm{\theta}_{i}^{t,s-2})}^2\\
    & \labelrel\leq{in_ld_divg_cahy} (1+\frac{1}{2\tau-1})\mathbb{E}_t\norm{\bm{\theta}_{i}^{t,s-2}-\bm{\theta}^t}^2
    +2\eta^2\tau\mathbb{E}_t\norm{\nabla f_i(\bm{\theta}_{i}^{t,s-2})-\nabla f_i(\bm{\theta}^{t})+\nabla f_i(\bm{\theta}^{t})}^2
    +\eta^2\zeta_i^2\\
    & \labelrel\leq{in_ld_divg_cahy1} (1+\frac{1}{2\tau-1})\mathbb{E}_t\norm{\bm{\theta}_{i}^{t,s-2}-\bm{\theta}^t}^2
    +4\eta^2\tau\mathbb{E}_t\norm{\nabla f_i(\bm{\theta}_{i}^{t,s-2})-\nabla f_i(\bm{\theta}^{t})}^2
    +4\eta^2\tau\mathbb{E}_t\norm{\nabla f_i(\bm{\theta}^{t})}^2
    +\eta^2\zeta_i^2\\
    & \labelrel\leq{in_ld_divg_smooth} (1+\frac{1}{2\tau-1})\mathbb{E}_t\norm{\bm{\theta}_{i}^{t,s-2}-\bm{\theta}^t}^2+4\eta^2L^2\tau\mathbb{E}_t\norm{\bm{\theta}_{i}^{t,s-2}-\bm{\theta}^{t}}^2+4\eta^2\tau\mathbb{E}_t\norm{\nabla f_i(\bm{\theta}^{t})}^2+\eta^2\zeta_i^2,
\end{align*}
where~\eqref{eq_ld_cross} holds due to the unbiased gradient estimation in Assumption~\ref{ass:gradient},~\eqref{in_ld_divg_cahy} follows from Assumption \ref{ass:gradient} and Lemma \ref{lemma_in_square_sum} with $\alpha=\frac{1}{2\tau-1}$,~\eqref{in_ld_divg_cahy1} follows from Lemma \ref{lemma_cau_sch_in}, and \eqref{in_ld_divg_smooth} follows from the smoothness assumption \ref{ass:smoothness}. Next, taking the average of sum of all clients, we get:
\begin{align*}
    \frac{1}{N}\sum_{i=1}^{N}\mathbb{E}_t&\norm{\bm{\theta}^t-\bm{\theta}_{i}^{t,s-1}}^2 \leq  (1+\frac{1}{2\tau-1}+4\eta^2L^2\tau)\frac{1}{N}\sum_{i=1}^{N}\mathbb{E}_t\norm{\bm{\theta}_{i}^{t,s-2}-\bm{\theta}^t}^2
    +\frac{4\eta^2\tau}{N}\sum_{i=1}^{N}\mathbb{E}_t\norm{\nabla f_i(\bm{\theta}^{t})}^2
    +\frac{\eta^2}{N}\sum_{i=1}^{N}\zeta_i^2\notag\\
    \labelrel\leq{in_ld_divg_dissm} & (1+\frac{1}{2\tau-1}+4\eta^2L^2\tau)\frac{1}{N}\sum_{i=1}^{N}\mathbb{E}_t\norm{\bm{\theta}_{i}^{t,s-2}-\bm{\theta}^t}^2+4\eta^2\tau\gamma^2\mathbb{E}_t\norm{\nabla f(\bm{\theta}^{t})}^2+4\eta^2\tau\kappa^2+\eta^2\bar{\zeta}^2,
\end{align*}
where \eqref{in_ld_divg_dissm} holds due to the dissimilarity Assumption \ref{ass:bound_dissimi} and the notation $\bar{\zeta}^2 = (1/N)\sum_{i=1}^{N}\zeta_i^2$. When $\eta\leq\frac{1}{2\sqrt{2}\tau L}$, then $4\eta^2L^2\tau\leq\frac{1}{2\tau}$. We have:
\begin{align*}
    \frac{1}{N}\sum_{i=1}^{N}\mathbb{E}_t\norm{\bm{\theta}^t-\bm{\theta}_{i}^{t,s-1}}^2 \leq & (1+\frac{1}{\tau-1})\frac{1}{N}\sum_{i=1}^{N}\mathbb{E}_t\norm{\bm{\theta}_{i}^{t,s-2}-\bm{\theta}^t}^2+4\eta^2\tau\gamma^2\mathbb{E}_t\norm{\nabla f(\bm{\theta}^{t})}^2+4\eta^2\tau\kappa^2+\eta^2\bar{\zeta}^2
\end{align*}
Unrolling the recursion, we get:
\begin{align*}
    \frac{1}{N}\sum_{i=1}^{N}\mathbb{E}_t\norm{\bm{\theta}^t-\bm{\theta}_{i}^{t,s-1}}^2 \leq & \sum_{h=0}^{s-1}(1+\frac{1}{\tau-1})^h\left[4\eta^2\tau\gamma^2\mathbb{E}_t\norm{\nabla f(\bm{\theta}^{t})}^2+4\eta^2\tau\kappa^2+\eta^2\bar{\zeta}^2\right]\\
    \leq & (\tau-1)\left[(1+\frac{1}{\tau-1})^\tau-1\right]\times\left[4\eta^2\tau\gamma^2\mathbb{E}_t\norm{\nabla f(\bm{\theta}^{t})}^2+4\eta^2\tau\kappa^2+\eta^2\bar{\zeta}^2\right]\\
    \labelrel\leq{in_ld_mods_tau} & 16\eta_{l}^2\tau^2\gamma^2\norm{\nabla f(\bm{\theta}^t)}^2 + 16\eta^2\tau^2\kappa^2 +4\tau\eta^2\bar{\zeta}^2,
\end{align*}
where~\eqref{in_ld_mods_tau} follows from $(1+\frac{1}{\tau-1})^\tau\leq5$ when $\tau>1$.
\end{IEEEproof}

\section{Useful Inequalities}
For ease of notation, we use $\|\cdot\|$ to denote the $\ell_2$ vector norm.
\setcounter{lemma}{5}
\begin{lemma}[Jensen's inequality]\label{lemma_jensen_in}
For arbitrary set of $N$ vectors $\{\bm{a}_i\}_i^N$, $\bm{a}_i\in\mathbb{R}^d$ and positive weights $\{w_i\}_{i\in [N]}$, $\sum_{i}^Nw_i=1$,
\begin{equation}
    \norm{\sum_{i=1}^N w_i \bm{a}_i}^2 \leq \sum_{i=1}^N w_i\norm{\bm{a}_i}^2
\end{equation}
\end{lemma}

\begin{lemma}[Cauchy-Schwarz inequality]\label{lemma_cau_sch_in}
For arbitrary set of $N$ vectors $\{\bm{a}_i\}_{i=1}^N$, $\bm{a}_i \in \mathbb{R}^d$,
\begin{equation}
    \norm{\sum_{i=1}^N\bm{a}_i}^2\leq N\sum_{i=1}^N\norm{\bm{a}_i}^2
\end{equation}
\end{lemma}

\begin{lemma}\label{lemma_in_square_sum}
For given two vectors $\bm{a},\bm{b}\in \mathbb{R}^d$,
\begin{equation}
    \norm{\bm{a}+\bm{b}}^2\leq(1+\alpha)\norm{\bm{a}}^2+(1+\alpha^{-1})\norm{\bm{b}}^2, \forall \alpha \geq 0.
\end{equation}
\end{lemma}

\begin{lemma}\label{lemma_in_inn_prod}
For given two vectors $\bm{a},\bm{b}\in \mathbb{R}^d$,
\begin{equation}
    2\langle \bm{a},\bm{b}\rangle \leq \pi\norm{\bm{a}}^2 +\pi^{-1}\norm{\bm{b}}^2, \forall \pi > 0.
\end{equation}
\end{lemma}

\section{Intermediate Results}

\begin{lemma}[Bounded Sparsification]\label{lemma_bound_spar}
Given a vector $\mathbf{a}\in \mathbb{R}^d$ and a parameter $k\in[d]$, The $\randk_k$ projection matrix $\mathbf{A}\in\mathbb{R}^{k\times d}$ generated from the activate subset $\omega$ holds that
\begin{gather}
    \mathbb{E}_{\omega}[\mathbf{A^{\intercal}Aa}]=\frac{k}{d}\mathbf{a}, \quad \mathbb{E}_{\omega}\norm{\mathbf{A^{\intercal}Aa}-\mathbf{a}}^2 = (1-\frac{k}{d})\norm{\mathbf{a}}^2.   
\end{gather}
\end{lemma}
\begin{IEEEproof}
For the $\randk_k$, taking the expectation on the active set $\omega$, we have:
\begin{align}
    \mathbb{E}_{\omega}[\mathbf{A^{\intercal}Aa}] = [\frac{k}{d}[\mathbf{a}]_1,\dots,\frac{k}{d}[\mathbf{a}]_d] = \frac{k}{d}\mathbf{a}
\end{align}
For the variance of $\randk_k$, we get:
\begin{align}
    \mathbb{E}_{\omega}\norm{\mathbf{A^{\intercal}Aa}-\mathbf{a}}^2 & = \sum_{n=1}^{d}\left(\frac{k}{d}([\mathbf{a}]_n-[\mathbf{a}]_n)^2+(1-\frac{k}{d})[\mathbf{a}]_n^2\right) = (1-\frac{k}{d})\norm{\mathbf{a}}^2
\end{align}
\end{IEEEproof}
\begin{lemma}[Bounded Local Model Update]\label{lemma_bound_loc_ups}
Under Assumptions \ref{ass:smoothness}, \ref{ass:gradient} and \ref{ass:bound_dissimi}, the local model update $\Delta_i^t$ at round $t$ is bounded by:
\begin{align}
    \frac{1}{N}\sum_{i=1}^N\mathbb{E}_t\norm{\frac{1}{\tau}\sum_{s=1}^{\tau}\mathbf{g}_{i}^{t,s-1}}^2
    \leq\frac{2L^2}{N\tau}\sum_{i=1}^N\sum_{s=1}^{\tau}\mathbb{E}_t\norm{\bm{\theta}^t-\bm{\theta}_i^{t,s-1}}^2 + 2(\gamma^2\norm{\nabla f(\bm{\theta}^t)}^2+\kappa^2) + \bar{\zeta}^2
\end{align}
where $\bar{\zeta}^2=(1/N)\sum_{i=1}^{N}\zeta_{i}^{2}$.
\end{lemma}
\begin{IEEEproof} 
\begin{align}
    \frac{1}{N}\sum_{i=1}^{N}\mathbb{E}_t\norm{\frac{1}{\tau}\sum_{s=1}^{\tau}\mathbf{g}_{i}^{t,s-1}}^2 
    & = \frac{1}{N}\sum_{i=1}^{N}\mathbb{E}_t\norm{\frac{1}{\tau}\sum_{s=1}^{\tau}\mathbf{g}_{i}^{t,s-1} - \frac{1}{\tau}\sum_{s=1}^{\tau}\nabla f_i(\bm{\theta}_{i}^{t,s-1}) + \frac{1}{\tau}\sum_{s=1}^{\tau}\nabla f_i(\bm{\theta}_{i}^{t,s-1})}^2\notag\\
    & \labelrel={eq_lmu_cross} \frac{1}{N}\sum_{i=1}^{N}\left[\mathbb{E}_t\norm{\frac{1}{\tau}\sum_{s=1}^{\tau}\mathbf{g}_{i}^{t,s-1}-\frac{1}{\tau}\sum_{s=1}^{\tau}\nabla f_i(\bm{\theta}_{i}^{t,s-1})}^2
    + \mathbb{E}_t\norm{\frac{1}{\tau}\sum_{s=1}^{\tau}\nabla f_i(\bm{\theta}_{i}^{t,s-1})}^2\right]\notag\\
    & = \frac{1}{N}\sum_{i=1}^{N}\left[\mathbb{E}_t\norm{\frac{1}{\tau}\sum_{s=1}^{\tau}\mathbf{g}_{i}^{t,s-1}-\frac{1}{\tau}\sum_{s=1}^{\tau}\nabla f_i(\bm{\theta}_{i}^{t,s-1})}^2
    + \mathbb{E}_t\norm{\frac{1}{\tau}\sum_{s=1}^{\tau}\nabla f_i(\bm{\theta}_{i}^{t,s-1})-\nabla f_i(\bm{\theta}^{t})+\nabla f_i(\bm{\theta}^{t})}^2\right]\notag\\
    & \labelrel\leq{in_lmu_cauchy_sch} \frac{1}{N}\sum_{i=1}^{N}\left[\mathbb{E}_t\norm{\frac{1}{\tau}\sum_{s=1}^{\tau}\mathbf{g}_{i}^{t,s-1}-\frac{1}{\tau}\sum_{s=1}^{\tau}\nabla f_i(\bm{\theta}_{i}^{t,s-1})}^2
    + 2\mathbb{E}_t\norm{\frac{1}{\tau}\sum_{s=1}^{\tau}\nabla f_i(\bm{\theta}_{i}^{t,s-1})-\nabla f_i(\bm{\theta}^{t})}^2\right]\notag\\ 
    &\quad + 2\frac{1}{N}\sum_{i=1}^{N}\mathbb{E}_t\norm{\nabla f_i(\bm{\theta}^{t})}^2\notag\\
    & \labelrel\leq{in_lmu_ass_bound_dissimi}\frac{1}{N}\sum_{i=1}^{N}\left[\mathbb{E}_t\norm{\frac{1}{\tau}\left(\sum_{s=1}^{\tau}\mathbf{g}_{i}^{t,s-1}-\sum_{s=1}^{\tau}\nabla f_i(\bm{\theta}_{i}^{t,s-1})\right)}^2
    + 2\mathbb{E}_t\norm{\frac{1}{\tau}\sum_{s=1}^{\tau}\nabla f_i(\bm{\theta}_{i}^{t,s-1})-\nabla f_i(\bm{\theta}^{t})}^2\right]\notag\\ 
    & \quad + 2(\gamma^2\norm{\nabla f(\bm{\theta}^t)}^2+\kappa^2)\label{in_lmu}
\end{align}
where~\eqref{eq_lmu_cross} follows from the unbiased stochastic gradient Assumption~\ref{ass:gradient},~\eqref{in_lmu_cauchy_sch} uses Lemma~\ref{lemma_cau_sch_in}, and~\eqref{in_lmu_ass_bound_dissimi} holds due to Assumption~\ref{ass:bound_dissimi}. 
\begin{align}
    \mathbb{E}_t\norm{\frac{1}{\tau}\left(\sum_{s=1}^{\tau}\mathbf{g}_{i}^{t,s-1}-\sum_{s=1}^{\tau}\nabla f_i(\bm{\theta}_{i}^{t,s-1})\right)}^2 \leq \frac{1}{\tau} \sum_{s=1}^{\tau} \mathbb{E}_t\norm{ \mathbf{g}_i^{t,s-1}-\nabla f_i(\bm{\theta}_{i}^{t,s-1}) }^2\leq \frac{1}{\tau} \sum_{s=1}^{\tau}\zeta_i^2 = \zeta_i^2, \label{in_sgd_gd_jens} 
\end{align}
where \eqref{in_sgd_gd_jens} follows from Lemma \ref{lemma_jensen_in} and Assumption \ref{ass:gradient}.
\begin{align}
   \mathbb{E}_t\norm{\frac{1}{\tau}\sum_{s=1}^{\tau}\left( \nabla f_i(\bm{\theta}_i^{t,s-1})-\nabla f_i(\bm{\theta}^t) \right)}^2 \leq \frac{1}{\tau}\sum_{s=1}^{\tau}\mathbb{E}_t\norm{\nabla f_i(\bm{\theta}_{i}^{t,s-1}) - \nabla f_i(\bm{\theta}^t)}^2 \leq \frac{L^2}{\tau}\sum_{s=1}^{\tau}\mathbb{E}_t\norm{\bm{\theta}_{i}^{t,s-1} - \bm{\theta}^t}^2,\label{in_grad_diff_jens}
\end{align}
where~\eqref{in_grad_diff_jens} follows from Lemma \ref{lemma_jensen_in} and Assumption \ref{ass:smoothness}. Combining~\eqref{in_lmu},~\eqref{in_sgd_gd_jens} and~\eqref{in_grad_diff_jens}, we have:
\begin{align}
     \frac{1}{N}\sum_{i=1}^{N}\mathbb{E}_t\norm{\frac{1}{\tau}\sum_{s=1}^{\tau}\mathbf{g}_{i}^{t,s-1}}^2
    \leq\frac{2L^2}{N\tau}\sum_{i=1}^N\sum_{s=1}^{\tau}\mathbb{E}_t\norm{\bm{\theta}^t-\bm{\theta}_i^{t,s-1}}^2+2(\gamma^2\norm{\nabla f(\bm{\theta}^t)}^2+\kappa^2)+\bar{\zeta}^2
\end{align}
\end{IEEEproof}

\section{Proof of Lemma 5}\label{append_proof_lemma_randk_powr}
\begin{IEEEproof}
Taking the expectation on the $\randk_k$, we have:
\begin{subequations}
\begin{align}
    \mathbb{E}\norm{\mathbf{A}^t\Delta_{i}^{t}}_{2}^{2} &= \sum_{j=1}^{d}\frac{k}{d}[\Delta_{i}^{t}]_{j}^{2}\\
    &=\frac{k}{d}\norm{\bm{\theta}_{i}^{t, \tau} - \bm{\theta}^{t}}_2^{2}\\
    &=\frac{k}{d}\eta^2\norm{\sum_{s=1}^{\tau} \mathbf{g}_{i}^{t,s-1}}_2^2\label{in_randk_norm}\\
    &\leq \frac{k}{d}\eta^2\tau^2 C_{1}^{2},\label{in_bound_grad}
\end{align}
\end{subequations}
where \eqref{in_randk_norm} holds due to \eqref{eq_round_sgd}, \eqref{in_bound_grad} follows from Assumption~\ref{ass_Bound_Grad} and Cauchy–Schwarz inequality.
\end{IEEEproof}

\end{document}